\documentclass[draftcls,onecolumn]{IEEEtran}
\usepackage{wrapfig} 
\usepackage{setspace,verbatim,amsfonts,graphicx,amsmath,amsthm,amsbsy,amssymb,epsfig,url}
\usepackage{amsthm}
\usepackage{gensymb}
\usepackage{graphicx}
\usepackage{hyperref}

\usepackage{cite}
\usepackage{makecell}

\usepackage{amsfonts}
\usepackage{amsmath,bm}
\usepackage{amssymb}
\usepackage{amsthm}
\usepackage{tikz,graphicx}
\usepackage{mathrsfs}
\usepackage{algorithmic}
\usepackage{algorithm}
\usepackage{array,multirow}
\usepackage{color}

\usepackage{mathdots}
\usepackage{MnSymbol}%
\usepackage{mathtools}
\usepackage{subfigure}

{}

\newfont{\smallmathfont}{cmmib10 at 8pt}

\newcommand{\Bb}{{\mathbf B}}
\newcommand{\Cb}{{\mathbf{C}}}
\newcommand{\Db}{{\mathbf D}}
\newcommand{\Eb}{{\mathbf E}}

\newcommand{\Hb}{{\mathbf H}}
\newcommand{\Ib}{{\mathbf I}}

\newcommand{\bb}{{\mathbf b}}

\newcommand{\kb}{{\mathbf k}}

\newcommand{\rb}{{\mathbf x}}

\newcommand{\mbf}{\mathbf}

\newcommand{\Phib}{{\boldsymbol {\Phi}}}
\newcommand{\Psib}{{\boldsymbol {\Psi}}}

\newcommand{\Sigmab}{{\boldsymbol {\Sigma}}}
\newcommand{\Rd}{{\mathbb R}}
\newcommand{\Cd}{{\mathbb C}}

\newcommand{\hank}{\mathbf{H}}

\newcommand{\beq}{\begin{equation}}
\newcommand{\eeq}{\end{equation}}
\newcommand{\beqa}{\begin{eqnarray}}
\newcommand{\eeqa}{\end{eqnarray}}

\newcommand{\Fc}{{\mathcal F}}
\newcommand{\rank}{\mathrm{rank}}
\newcommand{\norm}[1]{\left\lVert #1 \right\rVert}

\begin{document}


\title{Structured Low-Rank Algorithms: Theory, MR Applications, and Links to Machine Learning
}

\author{Mathews Jacob,~\IEEEmembership{Senior Member,~IEEE}, Merry P. Mani, and  Jong~Chul~Ye,~\IEEEmembership{Senior Member,~IEEE} 
\thanks{
		MJ and MM are with the University of Iowa, Iowa City, IA 52242 (e-mails: mathews-jacob@uiowa.edu,merry-mani@uiowa.edu). JCY is with the Department of Bio and Brain Engineering, Korea Advanced Institute of Science and Technology (KAIST), 
		Daejeon 34141, Republic of Korea (e-mail: jong.ye@kaist.ac.kr).
		}
\thanks{Corresponding author: Jong Chul Ye, Dept. of Bio and Brain Engineering,
Korea Advanced Inst. of Science \& Technology (KAIST)
291 Daehak-ro, Yuseong-gu, Daejeon 34141
Republic of Korea, Email: jong.ye@kaist.ac.kr
}
}

\maketitle

\vspace{-4em}
\begin{abstract}
In this  survey, we provide a detailed review of recent advances in the recovery of continuous domain multidimensional signals from their few non-uniform (multichannel) measurements using structured low-rank matrix completion formulation. This framework is centered on the fundamental duality between the compactness (e.g., sparsity) of the continuous signal and the rank of a structured matrix, whose entries are functions of the signal. This property enables the reformulation of the signal recovery as a low-rank structured matrix completion, which comes with performance guarantees. We will also review fast algorithms that are comparable in complexity to current compressed sensing methods, which enables the application of the framework to large-scale 
magnetic resonance (MR) recovery problems. The remarkable flexibility of the formulation can be used to exploit signal properties that are difficult to capture by current sparse and low-rank optimization strategies. We  demonstrate the utility of the framework in a wide range of MR imaging (MRI) applications, including highly accelerated imaging, calibration-free acquisition, MR artifact correction, and ungated dynamic MRI.
\end{abstract}

%

\vspace{-1em}\section{Introduction}

The slow nature of signal acquisition in magnetic resonance imaging (MRI), where the image is formed from a sequence of Fourier samples, often restricts the achievable spatial and temporal resolution in multi-dimensional static and dynamic imaging applications. Discrete compressed sensing (CS) methods provided a major breakthrough to accelerate the magnetic resonance (MR) signal acquisition by reducing the sampling burden.
{As described in an introductory article in this special issue \cite{doneva} these algorithms exploited the sparsity of the discrete signal  in a  transform domain to recover the images from a few measurements.



In this paper, we review a continuous domain extension of CS  using a structured low-rank (SLR) framework for the recovery of an image or a series of images from a few measurements using various compactness assumptions {\cite{barkhuijsen,liangslr,zhang,shin2014calibrationless,Uecker2014espirit,haldar2014low,qu2015accelerated,jin2016general,ongie2016off,cao,morasa,haldar2016p,ye2016compressive,ongie2017fast,ongie2017convex,jin2017mri,mani2017multi,bilgic,hu2018generalized,manifold1,haldarspm}.} 
The general strategy of the SLR framework starts with defining a \emph{lifting} operation to construct a structured matrix, whose entries are functions of the signal samples. The SLR algorithms exploit the dual relationships between the signal compactness properties (e.g. sparsity, smoothness) and the rank of the lifted matrix. This dual relationship allows recovery of the signal from a few samples in the measurement domain as an SLR optimization problem.  
While this strategy may seem contrived, the main benefit of this framework is its remarkable flexibility in exploiting the compactness properties of a variety of signal structures, including continuous-domain real-world signals that classical approaches have difficulty capturing.  For example, SLR approaches can recover continuous domain images from a few Fourier measurements with minimal discretization errors, unlike compressed sensing approaches. The SLR methods can account for signals with an infinite number of signal discontinuities that are localized to a curve/surface, which is more general than signals with a finite number of isolated signal discontinuities considered in recent super-resolution methods \cite{candessuper}. In addition, the SLR schemes come with fast algorithms that are readily applicable to large-scale imaging applications such as static and dynamic MRI, unlike super-resolution methods \cite{candessuper} that rely on semi-definite programming. 
  Another example of SLR methods is the recovery of images from their multichannel measurements with unknown sensitivities {\cite{shin2014calibrationless, jin2016general,haldar2016p,mani2017multi}.} These schemes rely on the low-rank structure of a structured matrix, obtained by concatenating block Hankel matrices, which are obtained from each channel image. It is difficult for classical convex compressed sensing algorithms to exploit such complex relations between multichannel measurements. The SLR schemes have also been recently extended to recover an ensemble of images that lie on a smooth surface in high-dimensional space. Current subspace model or a union of subspace model is not efficient in capturing the above property; inspired by kernel methods \cite{kernelbook}, one can define a non-linear mapping that will transform the smooth surface to a subspace. In SLR approaches, this structure can  be exploited by the construction of a structured matrix, whose columns are non-linearly transformed signal samples. 

The SLR framework is closely related to the
extensive work on linear prediction in MRI, which
was often formulated using structure low-rank matrix  \cite{Uecker2014espirit,haldar2014low,barkhuijsen,liangslr,zhang}. Early work in the context of MRI dates
back to 1985 in the context of MR spectroscopy
 \cite{barkhuijsen}, followed by its generalization to MRI by
modeling the images as piecewise 1D
polynomials by Liang et al. \cite{liangslr}.
 The finite rate of innovation (FRI) theory \cite{vetterli2002sampling} also considers the recovery of piecewise polynomials and similar models from a finite number of samples. Multichannel linear predictive relations were introduced in the parallel MRI context in \cite{zhang,Uecker2014espirit}, while linear-predictive k-space relations resulting from support and phase constraints were introduced in \cite{haldar2014low}. 
The readers are referred to \cite{haldarspm} in the same special issue, which is focused on a more elaborate review of linear prediction approaches and their history in the MRI context.  
	We will review recent {development in the low-rank structured matrix approaches for MRI}. 
	}
\begin{enumerate}
    \item The early 1-D approaches \cite{liangslr,vetterli2002sampling} are generalized to multidimensional continuous domain signals \cite{haldar2014low,jin2016general,ongie2016off}. Theoretical guarantees are also available for multidimensional signals, whose discontinuities are localized to sets of infinite size, but of zero Lebesgue measure (e.g., curves in 2D and surfaces in 3D) \cite{ongie2016off}.

    \item 
    Low-rank matrix recovery algorithms with
recovery guarantees are used to recover signals
from nonuniform measurements  \cite{jin2016general,ongie2017fast,haldar2014low}, as opposed
to the explicit annihilating filter estimation from the
uniform sampling setting in the classical FRI \cite{liangslr,vetterli2002sampling}. Note that an earlier work in this respect goes
back to the work by Dologlou et al \cite{dologlou1996mri}.
  \item 
  {The framework is generalized to the recovery of signals from non-uniform multichannel measurements with unknown channel responses, facilitating the reduction in samples and calibration-free recovery in a unified matrix completion framework {\cite{shin2014calibrationless,Uecker2014espirit,jin2016general,haldar2016p,mani2017multi}}}.
    \item The non-linear generalization of FRI theory enables the recovery of points on smooth surfaces in high-dimensional spaces, which facilitates the joint reconstruction of an ensemble of images from their few measurements \cite{manifold1}.
    \item 
    {The main practical benefit of SLR schemes is their ability to capture a broad range of signal priors, resulting in a wide range of applications}, including static MRI {\cite{shin2014calibrationless,haldar2014low,jin2016general,ongie2017convex,ongie2017fast}}, dynamic MRI, diffusion MRI {\cite{mani2017multi}}, 
   MR artifact correction \cite{jin2017mri}, 
    parallel MRI {\cite{shin2014calibrationless,Uecker2014espirit,jin2016general}}, 
    multi-contrast image recovery \cite{bilgic}, spectroscopic imaging \cite{qu,cao}, and field inhomogeneity compensation \cite{nguyen,arvindB0}. 
\end{enumerate}


The above generalizations come with theoretical recovery guarantees {\cite{ye2016compressive,ongie2017convex}}, and fast algorithms \cite{jin2016general,ongie2017fast}. In addition, this framework provides rich insights in the deep links between 1-D FRI sampling theory \cite{vetterli2002sampling}, CS \cite{CaRoTa06},  low-rank matrix completion, and super-resolution theory \cite{candessuper}.

\begin{figure*}[t!]
	\center
	\subfigure[SLR interpolation]{\includegraphics[width =0.4\textwidth]{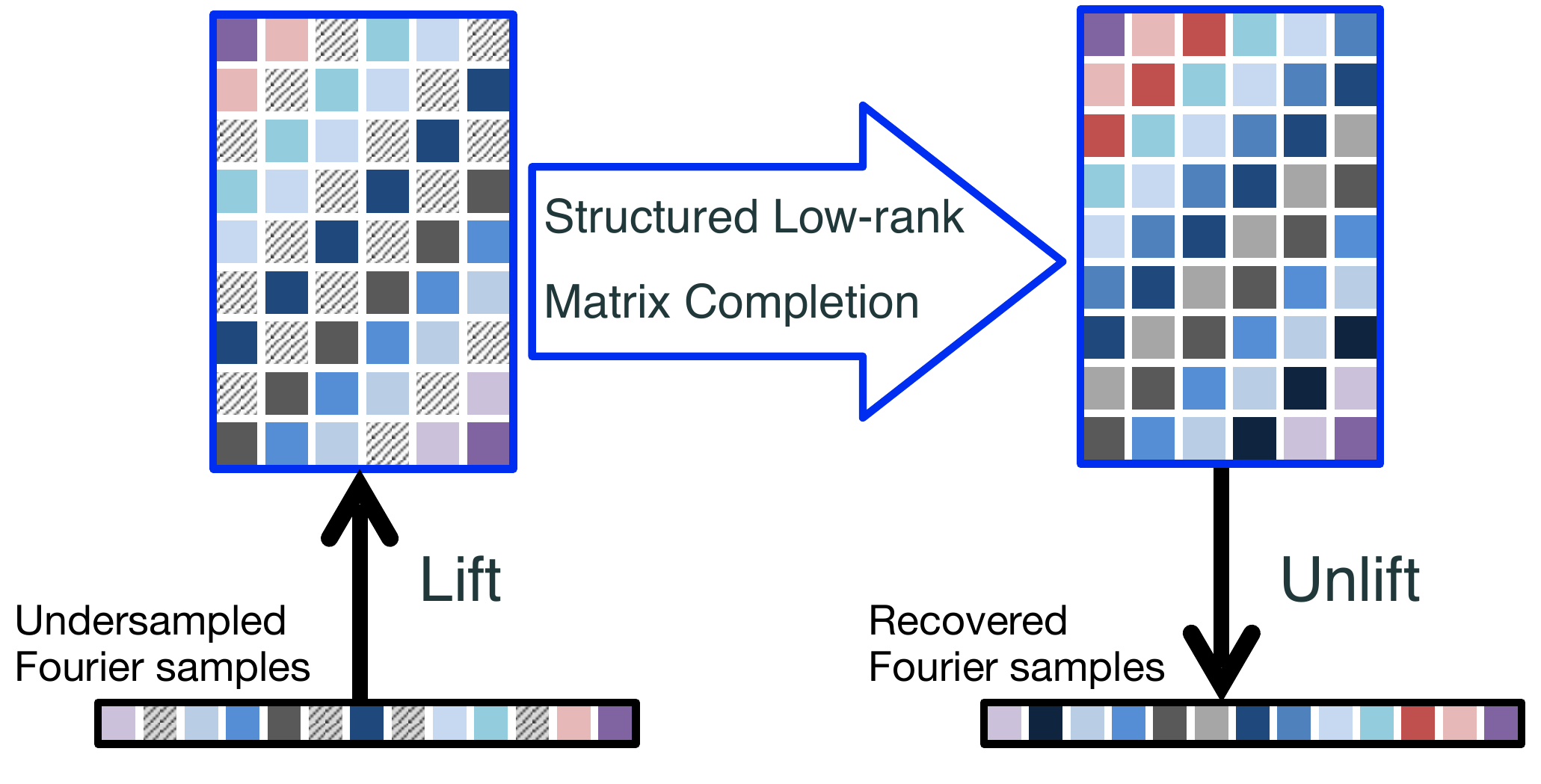}}~~
	\subfigure[SLR extrapolation]{\includegraphics[width =0.4\textwidth]{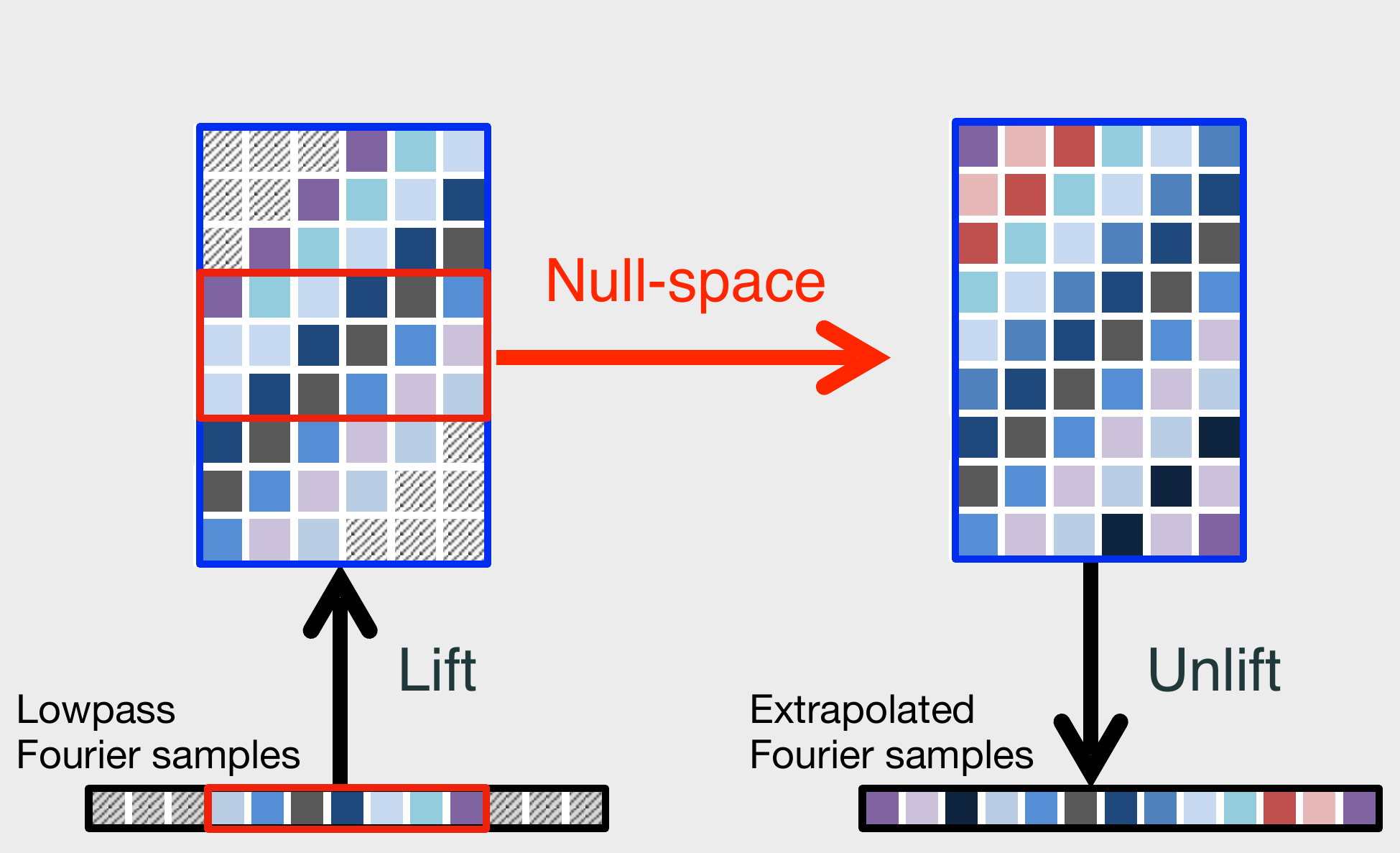}}~~
	\caption{Illustration of SLR-based interpolation and extrapolation methods in the context of 1-D FRI. (a) In SLR interpolation, the data is acquired on a non-uniformly sub-sampled Fourier grid. The SLR interpolation scheme relies on a lifting of the signal samples to a Hankel matrix, which has missing entries indicated by the hashed boxes. The one-to-one relation between the rank of a matrix and the continuous domain sparsity of the space domain signal is used to pose the recovery of missing samples as a structured low-rank matrix completion (SLRMC) problem in the lifted matrix domain. Specifically, the algorithm determines the matrix with the lowest rank that satisfies the Hankel structure and is consistent with the known matrix entries. Post-recovery, the matrix is unlifted to obtain the Fourier samples of the signal. (b) In SLR extrapolation problems, the low-frequency Fourier coefficients of the signal are uniformly sampled. The central fully known matrix region is used to estimate the null space of the matrix, which is used to linear-predict/extrapolate the missing high-frequency samples. The SLR algorithms that exploit the different signal structures  differ only in the structure of the lifted matrix; the algorithms are essentially the same. }
	\label{slrmc}
\end{figure*}

\vspace{-1em}\section{Overview}

\vspace{-1.3em}\subsection{Image acquisition in MRI}
In this section, we will briefly describe the image formation in MRI and introduce notations and terminologies used throughout the paper.

\subsubsection{Single-channel MRI measurements}
The  image acquisition in MRI constitutes the sampling of the Fourier transform of the image $f(\rb) $. The measurements in the Fourier domain (also referred to as k-space) are denoted by
\begin{eqnarray}\label{eq:coil}
\widehat f(\kb) = \left[\Fc f\right](\kb) := \int  f(\rb) e^{-i\,2\pi \kb^T \rb} d\rb. 
\end{eqnarray}
Here,  $\rb\in \Rd^d, d=2,\cdots,4$ and $\kb \in \mathbb Z^d$ denote the image domain and k-space coordinates, respectively.  
The goal of MR image recovery is then to reconstruct $f(\mathbf x)$ from the above measurements, which are measured on a {\em sparse} subset of the Fourier domain.  

\subsubsection{Multichannel measurements }
\label{multichannelmodel}
Modern scanners acquire the Fourier domain data using multiple receive coils to accelerate the acquisition. These receive coils have different spatial sensitivity patterns, thus providing complementary information. The measurement from the  $i$-th coil  is the Fourier transform of the coil-sensitivity weighted image and is denoted by
\begin{eqnarray}\label{eq:parallel}
\widehat f_i(\kb) = \left[\Fc f_i\right](\kb),\quad\mbox{where}~ f_i(\rb) =  s_i(\rb)f(\rb); ~i=1,\cdots, N_c. 
\end{eqnarray}
Here, $s_i(\rb) $ denotes coil sensitivities, $f_i(\rb) $ denotes the coil-sensitivity weighted image, and $N_c$ is the number of receive coils. The goal of parallel MR image (pMRI) recovery is then to reconstruct $f(\mathbf x)$ from a  few measurements with or without the knowledge of the coil sensitivities.

While the term {\it multichannel measurements} typically implies the multi-coil measurements described above, we will use the term in a broader sense. As discussed later, different k-space regions are often acquired with slightly different acquisition conditions in MRI. The distortions can often be modeled as spatial weighting terms, similar to coil sensitivities. For example, the signals corresponding to the odd-only and the even-only lines of k-space often differ in phase errors; we consider it  a two-channel acquisition with unknown spatial weighting terms.  

\vspace{-1.3em}\subsection{Structured Low-Rank methods: Bird's-eye view}
The SLR framework offers a versatile toolbox that exploits a variety of properties of continuous domain multichannel signals without the need for discretization. The SLR algorithms rely on a \emph{lifting} of the original signal to a matrix (see Fig. \ref{slrmc}); the structure of the matrix depends on the specific signal properties (e.g., continuous domain sparsity, multichannel relations). The framework relies on the duality relations between the \emph{compactness} of the signal (e.g., sparsity) and the \emph{rank} of the lifted matrix. 

{\it Interpolation via structured low-rank matrix completion}: In several accelerated MRI acquisitions, the measurements from the full k-space locations are not available.    
When the signal samples are acquired in a non-uniform fashion, one can rely on a structured low-rank matrix completion to interpolate the missing entries in the lifted matrix. Specifically, this entails determining a matrix with the lowest rank that preserves the structure of the lifted matrix and is consistent with the measured matrix entries. Once the matrix is completed, one can perform  inverse lifting to recover the samples of the continuous domain signal (see Fig. \ref{slrmc}.(a)). We term this class of methods as interpolation schemes{\cite{shin2014calibrationless,jin2016general,ongie2017fast,haldar2014low}}. 

{\it Extrapolation using two-step SLR algorithms} : In some applications, a certain limited region of k-space is acquired without any under-sampling (usually referred to as the calibration region, typically in the low-frequency regions). The super-resolution approaches in signal processing \cite{candessuper} aim to extrapolate these Fourier coefficients to high-frequency regions, thus recovering the images at higher resolution. The SLR extrapolation strategy is to estimate the null space of the lifted matrix from the fully sampled rows (indicated by the red boxes in Fig. \ref{slrmc}.(b)). 
Given the null space filters, one can (i) estimate a signal model based on the roots of the null-space filters, followed by linear estimation of the unknown signal model parameters, or (ii) perform linear prediction to extrapolate the high-frequency samples from the measured ones subject to the null-space constraints{\cite{ongie2016off,liangslr,zhang}}.

\vspace{-1em}\section{Signal Extrapolation using SLR}
\label{extrapolationsection}
We will now focus on specific continuous domain signal models and reveal their connection to the lifted matrices that facilitates the recovery of the continuous domain signal.  In this section, we will focus on SLR extrapolation where the low-frequency Fourier region is fully sampled as in Fig. \ref{slrmc}.(b).

\vspace{-1.3em}\subsection{FRI theory for piecewise smooth 1D signals }
\label{1dfri}
\begin{figure*}[t!]
 	\center
 	\subfigure[1-D FRI scheme]{\includegraphics[width =0.4\textwidth]{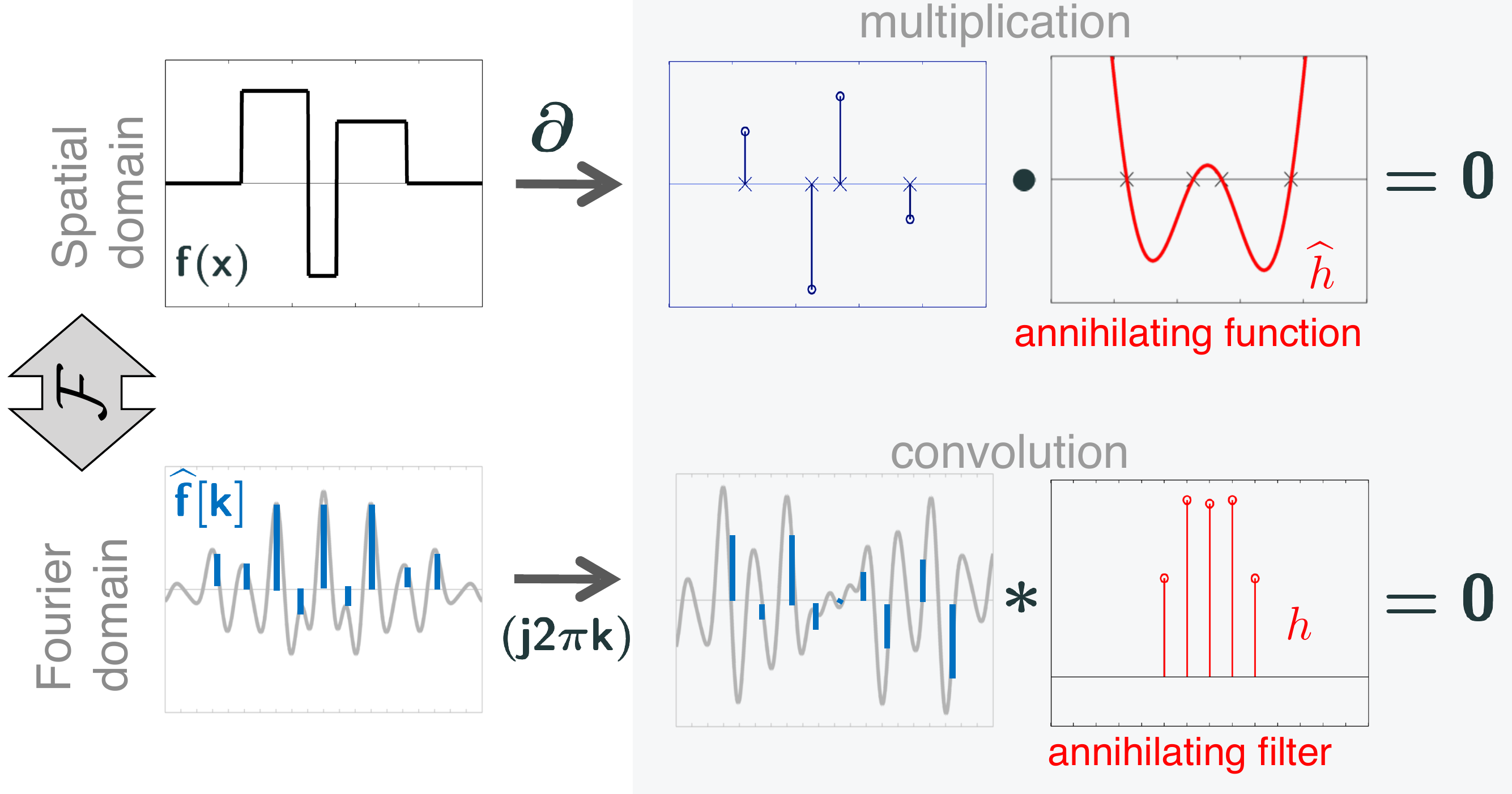}}~~~~
 	\subfigure[Extension of FRI to 2D]{	\includegraphics[width =0.4\textwidth]{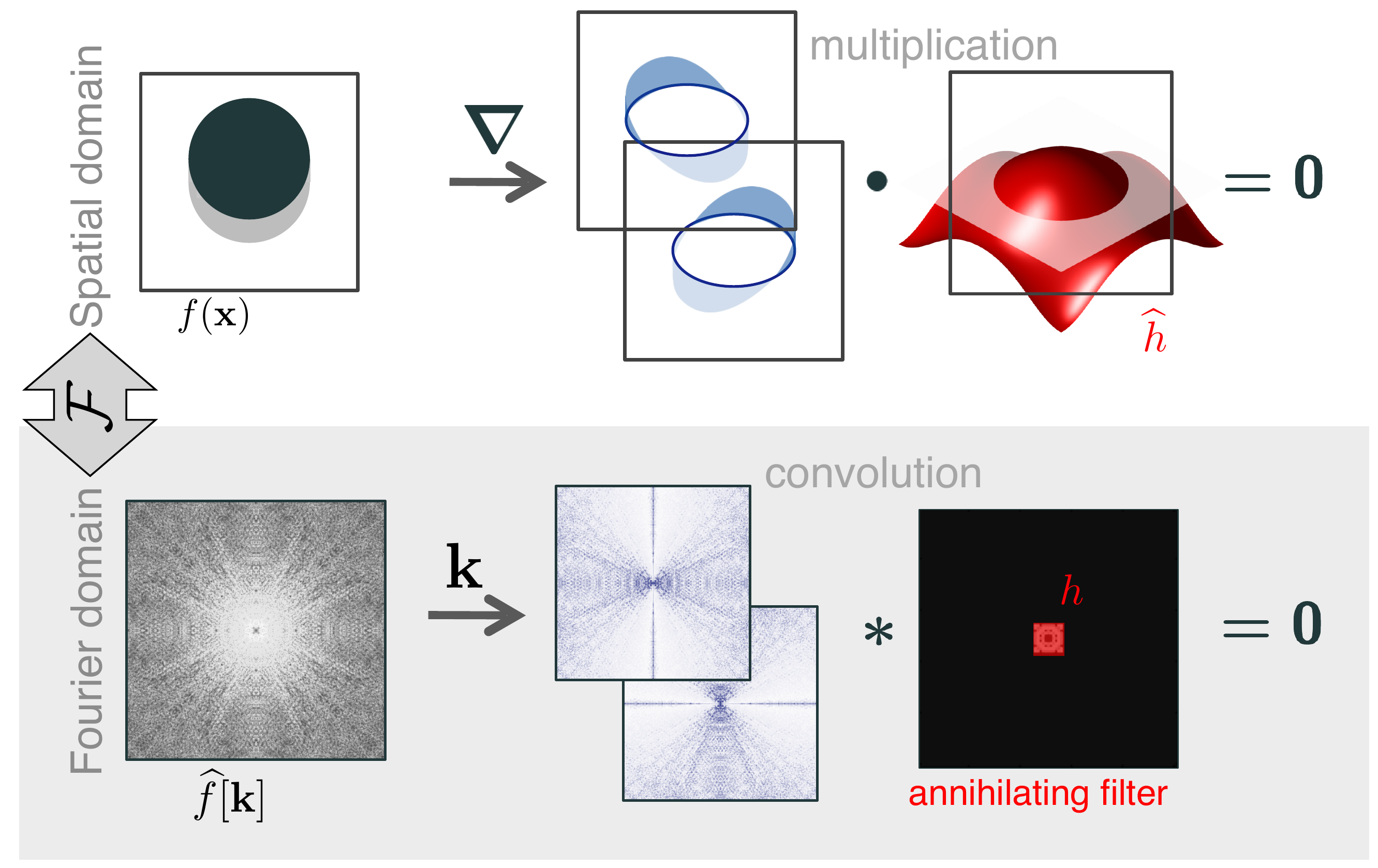}}
 	\caption{Illustration of 1-D and 2-D FRI relations. (a) We illustrate Prony's results in the light blue box, which demonstrates the recovery of Diracs at arbitrary locations. The Fourier coefficients of this signal consist of complex exponentials denoted by $y_k$ in the bottom row. Prony's results show that complex exponentials can be annihilated by the convolution with a filter $h$. This Fourier domain convolution relation can also be viewed as a multiplication-based annihilation relation in the space domain. These results can be extended to the recovery of piecewise constant 1-D signals by considering the derivative. Specifically, the derivative of piecewise constant signals consists of a linear combination of Diracs, which brings the problem to Prony's setting. (b). Extension of 1-D FRI to the 2-D setting. The gradient of the piecewise constant signal consists of a series of Diracs supported on curves and the location of image edges. We observe that the gradient can be annihilated by the multiplication by a function $\hat h$, whose zeros overlap with the image edges. If the $\hat h$ is bandlimited, the convolution with the Fourier coefficients of the image with the filter coefficients will also be zero. We generalize the notion of 1-D sparsity in FRI with the bandwidth of $\hat h$; a more bandlimited $\hat h$ will correspond to smoother curves, as shown in \cite{ongie2016off}. The above framework can be used to recover continuous domain piecewise constant images, whose edges are localized to bandlimited zero-measure curves of infinite support. This simple piecewise constant model is extended to more general piecewise smooth models in \cite{hu2018generalized}.}\vspace{-2em}
 	\label{Fig1}
 \end{figure*}

This approach \cite{barkhuijsen,liangslr,vetterli2002sampling} is a generalization of Prony's method to continuous domain 1-D signals. Consider the recovery of $r$-stream of Diracs $f$ (see Fig. \ref{Fig1}.(a)) at locations $x_j, j=1,..,r$ with weights $w_j$:
\begin{equation}\label{eq:signal3}
f(x) = \sum_{j=1}^{r} ~w_j ~\delta \left( x- x_j \right) \ , \quad x_j \in [0, 1].
\end{equation}
The discrete Fourier transform of the above continuous domain signal is a linear combination of complex exponentials with frequencies $\alpha_j = e^{-i2\pi x_j }$, specified by  
\begin{eqnarray}\label{eq:exp}
\widehat{ f} [k] = \sum_{j=1}^{r}~ w_j ~(\alpha_j)^k, \quad \forall k \ .
\end{eqnarray}
The classical Prony's results \cite{vetterli2002sampling} rely on the annihilation of such exponential signals by 
\begin{eqnarray}\label{eq:afilter}
\widehat h(z)  = \prod_{j=1}^{r} ~\left(1- \alpha_j z^{-1}\right) &=&  \sum_{n=0}^r  h[n]z^{-n} ,
\end{eqnarray}
such that the associated
$(r+1)$-tap filter $\widehat h[n]$  annihilates the Fourier samples $\widehat f$: 
\begin{eqnarray}\label{eq:annihilating}
( h\ast\widehat f )[k] &=& 0, 
\quad \forall k
\end{eqnarray}
This linear convolution relation \eqref{eq:annihilating} can be re-expressed in the matrix form (see Fig. \ref{slrmc}) to solve for $\widehat h$:
\begin{eqnarray}\label{eq:X2}
\underbrace{\left[
	\begin{array}{cccc}
	\widehat f[0]  &\widehat f[1] & \cdots   &\widehat f[r]   \\
	\widehat f[1]  &\widehat f[2] & \cdots &  \widehat f[r+1] \\
	\vdots    & \vdots     &  \ddots    & \vdots    \\
	\widehat f[r-1]  &\ldots & \cdots &\widehat f[2r-1]\\
	\end{array}
	\right]}_{\hank_{[r+1]}^{[2r] }(\widehat f)} \underbrace{\begin{bmatrix} h[r] \\  \vdots\\  h[0] \end{bmatrix}}_{\mathbf h} = 0   \  .
\end{eqnarray}
Here, the notation $\hank_{[r+1]}^{[2r] }(\widehat f)$ represents the linear convolution matrix,  
which is defined for the 2r samples of $\widehat f$ to be convolved with a filter of size $r+1$; here $[r]$ denotes the set $0,1,..,r$. Note that $\hank_{[r+1]}^{[2r] }(\widehat f)$ is a Hankel structured matrix and that the {\it matrix lifting} thus originates from the linear convolution embedded in \eqref{eq:annihilating}-\eqref{eq:X2} and is illustrated in Fig. \ref{slrmc}. The relation \eqref{eq:X2} can be used to identify $\mathbf h$. Once this linear-prediction filter $\mathbf h$ is available, the linear predictive relations \eqref{eq:annihilating} can be used to extrapolate the signal samples $\widehat f[k]; k=0,..,2r-1$ to any extent in Fourier space. 


\subsubsection{Theoretical Guarantees}
Based on \eqref{eq:X2}, the filter $\mathbf h$ can be estimated as a null-space vector of the {\it lifted} matrix as shown in Fig. \ref{slrmc}(b). If $2r$ signal consecutive samples are available, the null-space vector is unique \cite{liangslr,vetterli2002sampling}. 
Note from \eqref{eq:afilter} that the roots/zeros of $\widehat h(z)$ will specify  $\alpha_j$ in \eqref{eq:exp}. One can then solve a system of equations with $r$ unknowns to recover the  $w_j$ in \eqref{eq:signal3}; this estimation is unique with $2r$ measurements. 
 
\subsubsection{Spatial annihilation relations} The above results can be generalized to recover piecewise constant signal $f$ (see Fig. \ref{Fig1}.(a)), with $r$ discontinuities located at $t_j; j=1,..,r$. The derivative of the signal, denoted by $\partial f$, is a periodic stream of Diracs as in \eqref{eq:signal3}. The above theory can be adapted to this setting by considering the Fourier coefficients of the derivative of $f(x)$ as illustrated in Fig. \ref{Fig1}(a). The Fourier domain convolution based annihilation relations can also be viewed as space domain multiplication-based annihilation relation 
 \begin{equation}\label{space}
 \partial f(x) \cdot \hat h(x)=0
 \end{equation}
as shown in Fig. \ref{Fig1}(a). 
{Here, the space domain function $\hat h(x)$ is the Fourier transform of the filter $h[n]$ that annihilates the Fourier coefficients of $f$. } The zeros of $\widehat{h}$ overlap with the location of Diracs or the location of the discontinuities of the piecewise constant signal{\cite{liangslr}}. We will now use this spatial domain annihilation relations to extend the 1-D results to higher dimensions.
  
\vspace{-1.3em}\subsection{Piecewise smooth signals in higher dimensions}
\label{pwcsection}

We now review  how to extend the classical 1-D FRI theory to recover the piecewise constant images $f(\mathbf x)$ shown in Fig. \ref{Fig1}(b). We assume that the Fourier samples are available in a rectangular region $\Gamma \subset \mathbb Z^2$ \cite{ongie2016off}. Note that the edges of $f$ consist of a set of curves, denoted by $\mathcal C$. Extending the space domain annihilation relation in \eqref{space}, we assume the edge locations of the image to be represented by the zero level-sets of a 2-D bandlimited function $\widehat h$   \cite{ongie2016off} :
\begin{equation}\label{levelset}
\mathcal C = \{\mathbf x~| ~\widehat h(\mathbf x) =0\}
\end{equation}
where
\begin{equation}\label{bl}
\widehat h(\mathbf x)= \sum_{\mathbf k\in \Lambda}  h[\mathbf k] \,\exp(i\mathbf k^T \mathbf x)
\end{equation} is bandlimited to a rectangular region $\Lambda \subset \mathbb Z^2$. The bandwidth of $\widehat h$, denoted by $|\Lambda|$, is a measure of the complexity of the edge set \cite{ongie2016off,pan2014sampling}, which generalizes the notion of sparsity in compressive sensing. 

\begin{figure}[t!]
	\center
	\subfigure[Vertical Stacking]{\includegraphics[width =0.34\textwidth]{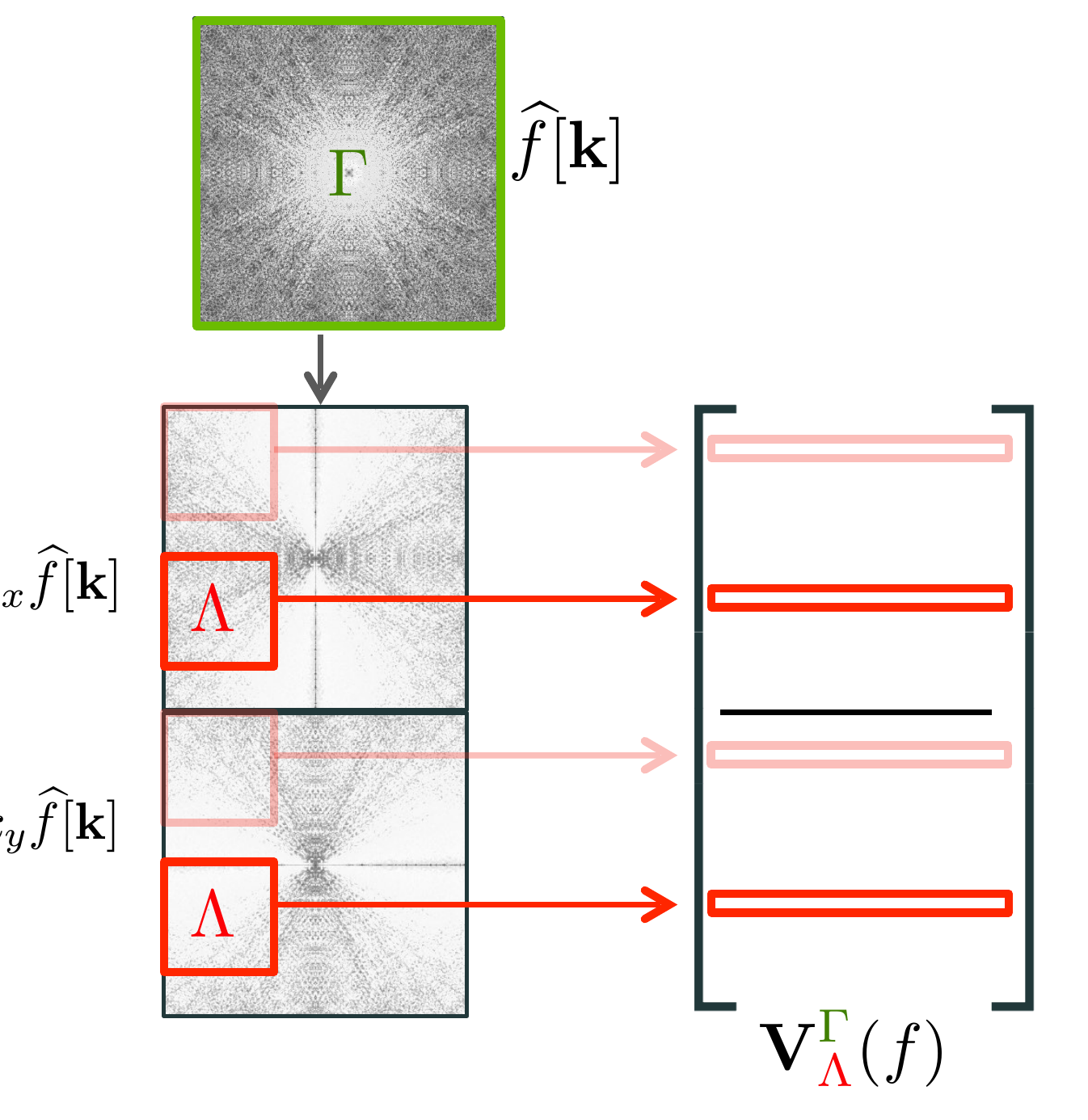}}
	\subfigure[Linear Stacking]{	\includegraphics[width =0.33\textwidth]{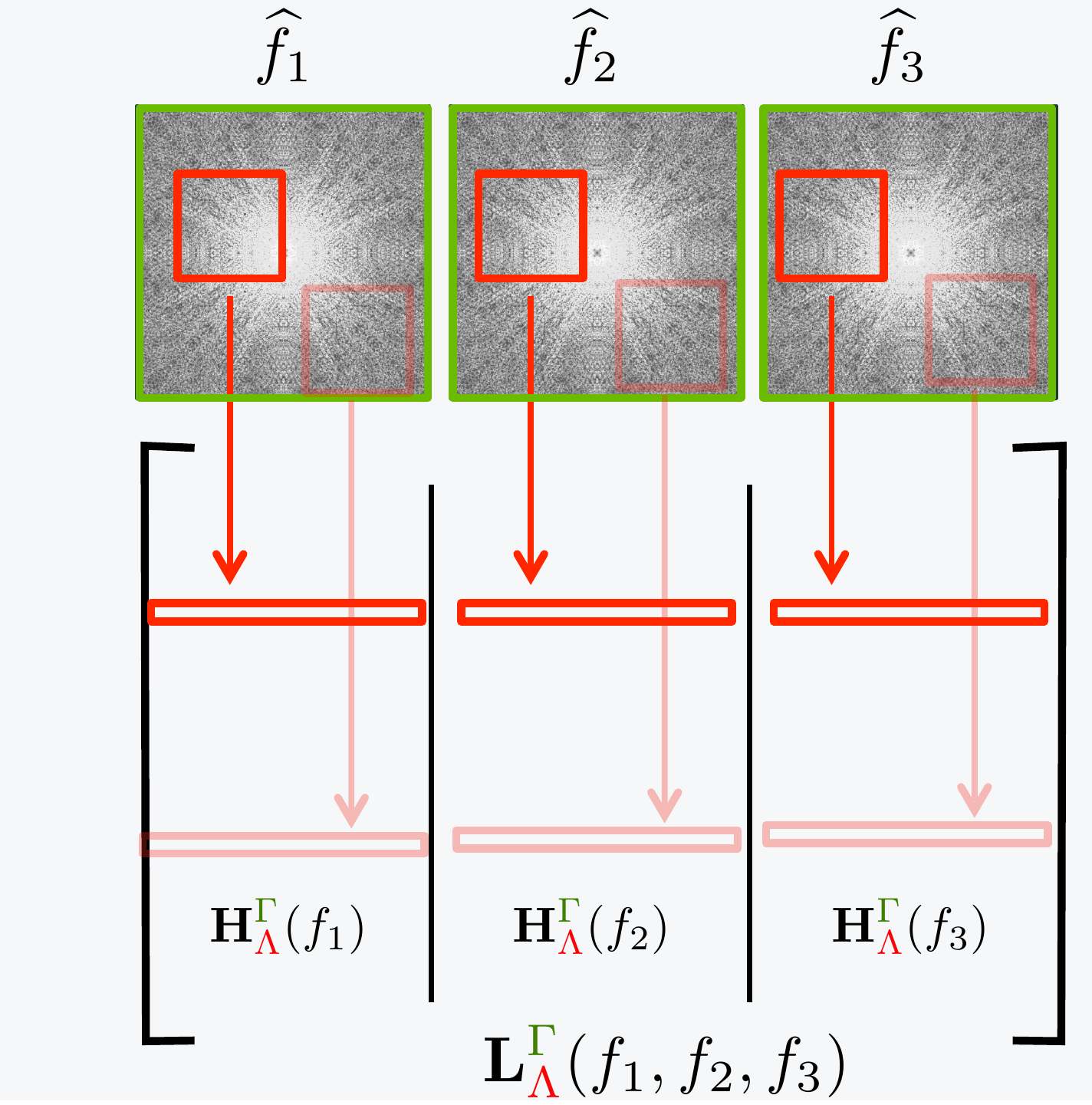}}
	\caption{Illustration of the matrix lifting operations in 2-D. (a). Vertical stacking of block Hankel matrices in  \eqref{lifted}, which is the matrix form of the convolutional relations in \eqref{eq:annsys}. The block Hankel matrices $\mathbf H_{\Lambda}^{\Gamma}(i\omega_x\hat f)$ are constructed from the Fourier coefficients of $i\omega_x\hat f$. Each row of  $\mathbf H_{\Lambda}^{\Gamma}(i\omega_x\hat f)$ corresponds to a patch of size $\Lambda$, drawn from the the Fourier coefficients supported in $\Gamma$. The number of valid shifts of $\Lambda$ within $\Gamma$ is denoted by $\Gamma \ominus \Lambda$ in  \eqref{eq:annsys}. (b) Linear stacking of block Hankel matrices in \eqref{hankmultichannel}. The block Hankel matrices of the different channels $f_i$ are concatenated in a linear fashion to construct the composite lifted matrix. When the low-frequency Fourier samples are fully acquired, the corresponding rows of the matrices are fully available. This facilitates the estimation of the null-space from these regions, which can be used to recover the remaining k-space regions as described in Section \ref{extrapolationsection}. When the Fourier grid is sub-sampled, the matrices can be completed using nuclear norm minimization as discussed in Section \ref{interpolation}. }\vspace{-2em}
	\label{Figstacking}
\end{figure}

Similar to \eqref{space}, we have space domain annihilation relation (see Fig. \ref{Fig1}(b)) specified by $
\nabla f \cdot \widehat h=0$; the Fourier transform of this relation provides the 2-D Fourier domain annihilation relations similar to \eqref{eq:annihilating}:
\begin{equation}
\sum_{\mathbf k \in {\Lambda}}  \widehat{\nabla f}[\boldsymbol\ell - \mathbf k] \;h[\mathbf k]= \boldsymbol 0, 
~~\forall ~\boldsymbol \ell\in \Gamma\ominus\Lambda.
\label{eq:annsys}
\end{equation}
Here, $\Gamma$ is the set of Fourier measurements of $f$. 
{Note that the convolution of $\widehat {\nabla f}$ with $h$ at a location $\mathbf k$ requires the samples of $\widehat {\nabla f}$ within the set $\mathbf k+\Lambda$, where the addition amounts to the translation of the set $\Lambda$ to the location $\mathbf k$. Since we only have the Fourier samples of $f$ within $\Gamma$, the convolutions specified by \eqref{eq:annihilating} can only be evaluated  within the set $\Gamma\ominus \Lambda$, which is the morphological erosion of the set $\Gamma$ by $\Lambda$. } Similar to \eqref{eq:X2}, the convolution relation in \eqref{eq:annsys} can be rewritten in the matrix form similar to \eqref{eq:X2} as 
\begin{equation}\label{lifted}
\underbrace{\begin{bmatrix}\mathbf H_{\Lambda}^{\Gamma}(i\omega_x \widehat f)\\\mathbf H_{\Lambda}^{\Gamma}(i\omega_y \widehat f)\end{bmatrix}}_{\mathbf V_{\Lambda}^{\Gamma}\left(\widehat{f}\right)} \;\mathbf h = 0,
\end{equation}
where $\mathbf h$ represents the vectorized filter coefficients. Here,  $\mathbf H_{\Lambda}^{\Gamma}(i\omega_x \widehat f)$ is the matrix corresponding to the 2-D convolution of $i\omega_x \widehat f$, and  $\mathbf V_{\Lambda}^{\Gamma}\left(\widehat{f}\right)$ is a composite matrix obtained by stacking the block Hankel matrices vertically as shown in Fig. \ref{Figstacking}.(a). 

\subsubsection{Theoretical Guarantees} The filter $\mathbf h$, and equivalently the edges of $f$ can be estimated from \eqref{lifted} as the null-space vector of $\mathbf V_{\Lambda}^{\Gamma}\left(\widehat{ f}\right)$. It is shown in \cite[Theorem 1]{ongie2016off} that $\mathbf V_{\Lambda}^{\Gamma}(\widehat f)$ will have a unique null-space vector when the uniform Fourier samples of $\widehat f$ are available within $3\Lambda$, which is the three-fold dilation of $\Lambda$. The piecewise constant signal will then be uniquely identified from its Fourier measurements within $3\Lambda$ \cite[Theorem 4]{ongie2016off}. 

Note that the piecewise constant signal model in this section assumes a finite or infinite number of signal discontinuities, supported on a set of zero measure. This framework is more general than \cite{candessuper}, which assumes a finite number of isolated discontinuities. 
We note that even more general continuous
domain models \cite{haldar2014low}, which assume the signal
support to be a region of non-zero measure, do
exist. Since such models have infinite degrees of
freedom, it is difficult to guarantee the recovery of
such a signal from finite measurements.

\vspace{-1.3em}\subsection{Exponential 1-D signals} 
\label{exponentials}
Recovery of exponentially decaying signals are of interest in MRI since the temporal evolution of MRI signals in the presence of chemical shifts, field inhomogeneity and multiple relaxation mechanisms can be modeled as a linear combination of exponentials \cite{barkhuijsen,qu,cao}. Specifically, the 1-D exponential signal at the spatial location $\mathbf x$ evolving as a function of time can be expressed as

\begin{equation}\label{key}
f(\mathbf x,t) = \sum_{j=1}^r a_j(\mathbf x) ~\left(\beta_j(\mathbf x)\right)^t; ~~t = 1,.., T, 
\end{equation}
where the goal is to estimate the exponential parameters from their time series. Since the relation in Eq. \eqref{eq:afilter} is true for the above signal model, one can build Hankel matrices $\Hb_{[r]}^{[n]}(\widehat {f_{\mathbf x}})$ of the form \eqref{eq:X2} of each pixel at location $\mathbf x$, analogous to the description in Section \ref{1dfri}. Following \eqref{eq:exp}-\eqref{eq:X2}, these matrices will have a null space vector parameterized by the exponential parameters.  The roots of the null space vector are estimated using singular value decomposition (SVD), which are used to identify the exponential parameters. These sub-space strategies based on SVD of $\mathbf H_{[r]}^{[n]}(\widehat f_{\mathbf x})$ is widely used in the context of MR spectroscopy \cite{barkhuijsen,qu,cao}. 

\vspace{-1em}\section{Signal interpolation using SLR}
\label{interpolation}
This section shows how the extrapolation-based approaches in the previous section can be extended to the non-uniform setting. As discussed in Fig. \ref{slrmc}, we will exploit the duality between the rank of the lifted matrices and the compactness priors to fill in the missing entries. Note that such sampling patterns are of high relevance in CS applications. The generalization of SLR can be acheived with the theoretical performance guarantees that is similar to that of the standard CS approaches \cite{CaRoTa06}.

\vspace{-1.3em}\subsection{Piecewise smooth 1-D signals}
\label{1dtheory}
Consider the signal model in Section \ref{1dfri} with $r$ Diracs. Consider a Hankel structured matrix $\hank_{[d]}^{[n]}(\widehat f) \in \Cd^{(n-d)\times d}$ with $d>r$. Then, the authors in \cite{ye2016compressive} showed that
\begin{eqnarray}\label{eq:rankr}
\rank \left(\hank_{[d]}^{[n]}(\widehat f)\right)=  r .  
\end{eqnarray} 
{Similar observation was made in early work  \cite{liangslr}}.
The authors of  \cite{ye2016compressive}  further showed that \eqref{eq:rankr}  holds for general FRI signals with the minimum annihilating filter size of $r + 1$ on its spectral domain.
This property is useful for recovering sparse signals from Fourier measurements.
Specifically, let $\chi$ be a multi-set consisting of random indices
from $\{0,\ldots,n-1\}$ such that $|\chi|=m < n$.
Then, as shown in Fig. \ref{slrmc}.(a), we can  interpolate $\widehat f[k]$ for all $k\in \{0,\ldots, n-1\}$ from the sparse Fourier samples exploiting \eqref{eq:rankr}: 
\begin{equation}
\label{eq:nucmin}
\underset{\widehat g \in \mathbb{C}^n}{\text{minimize}} ~\norm{\hank_{[d]}^{[n]}(\widehat g)}_* 
\text{subject to} ~P_\chi (\widehat g) = P_{\chi} (\widehat f).
\end{equation}
where $\|\cdot\|_*$ denotes the matrix nuclear norm. Here, $P_{\chi}$ denotes a projection operation to the set $\chi$.
{Instead of using the nuclear norm, a similar problem was solved by nonconvex optimization method \cite{haldar2014low,dologlou1996mri}}.
%

\subsubsection{Theoretical Guarantees} If the sampling pattern  satisfies the incoherence condition with the parameter $\rho$,  then there exists a constant $c$ such that
$\widehat f$ is the unique minimizer to \eqref{eq:nucmin} with probability $1 - 1/n^2$\cite{ye2016compressive}, provided
\begin{equation}
\label{eq:samp_comp}
|\chi| \geq c~ \rho  ~ r ~\log^\alpha n,
\end{equation}
where $\alpha = 2$ if the discontinuities are located on uniform grid; $\alpha = 4$ for general continuous domain signals.
This suggests a near-optimal
performance similar to the discrete-domain CS approaches \cite{CaRoTa06}.

\vspace{-1.3em}\subsection{Piecewise smooth multidimensional signals}
\label{2dtheory}

We now consider the recovery of the Fourier samples of the piecewise constant image described in Section \ref{pwcsection} on $\Gamma \subset \mathbb Z^2$. We  consider matrices $\mathbf V_{\Lambda_1}^{\Gamma}\left(\widehat{ f}\right)$ shown in \eqref{lifted}, where $\Lambda\subset \Lambda_1$. It is shown in \cite{ongie2016off} that 
\begin{equation}\label{rankbound}
r=\rank ~\big(\mathbf V_{\Lambda_1}^{\Gamma}\left(\widehat{f}\right)\big) ~~=~~ |\Lambda_1|-|\Lambda_1\ominus \Lambda|.
\end{equation}
It is also shown in \cite{ongie2016off} that there is a \emph{duality} relation between the above rank and the smoothness of the edge set of the images. The low-rank structure of $\mathbf V_{\Lambda_1}^{\Gamma}$ enables the recovery of $\widehat f$ on $\Gamma$ using SLR. 

\subsubsection{Theoretical Guarantees} The results in \cite[Theorem 4]{ongie2017convex} show that the SLR algorithm recovers the true Fourier samples $\widehat f[\mathbf k]; \mathbf k \in \Gamma $ from its uniform random measurements on $\chi \subset \Gamma$ with probability exceeding $1-|\Gamma|^{-2}$, provided
\begin{equation}\label{probrec}
|\chi| \geq c~\rho~c_s~ r~ \log^4 |\Gamma|,
\end{equation}
where $r$ is defined as in \eqref{rankbound}, $c_s = |\Gamma|/|\Lambda_1|$, $c$ is a universal constant, and $\rho>1$ is an incoherence measure depending on the geometry of the edge set defined in \cite{ongie2017convex}. Note that the result bears remarkable similarity to \eqref{eq:samp_comp}, in the case of continuous domain signals not on the grid.

\vspace{-1.3em}\subsection{Exponential signals with spatially smooth parameters}
\label{expslr}
The approaches in Section \ref{exponentials} considered the pixel-by-pixel recovery of exponential parameters at each spatial location $\mathbf x$ from their uniform samples. The recovery of exponentials from non-uniform $k-t$ domain samples is key to accelerating the acquisition in several MR applications. Note that the coefficients of the annihilation filters $h_{\mathbf x}[z]$ depend on the exponential parameters at the specific pixel location $\mathbf x$. In imaging applications, the exponential parameters are often spatially smooth. The spatial Fourier transform of the filter coefficients can be assumed to be support limited to a 3-D cube of size $\Lambda$. The annihilation relations thus can be compactly expressed as 
\begin{equation}
\mathbf H_{\Lambda}^{\Gamma}(\widehat f) ~\mathbf h= 0,
\end{equation}
where $\mathbf H_{\Lambda}^{\Gamma}(\widehat f)$ is the block Hankel matrix corresponding to the 3-D convolution of $\widehat f[\mathbf k,n]$ with the filter $\mathbf h$. Similar to the discussions in Sections \ref{1dtheory} and \ref{2dtheory}, the matrix $\mathbf H_{\Lambda_1}^{\Gamma}(\widehat f)$ is low-rank if $\Lambda \subset \Lambda_1$. Thus, the recovery of the 3-D data set from its undersampled k-t space measurements can be posed as a structured low-rank problem.

\vspace{-1em}\section{Recovery of multichannel MRI data}
We now consider the recovery of multichannel data using calibration-based and calibration-free  strategies, which are analogous to the extrapolation and interpolation methods discussed above.
 \vspace{-1.3em}\subsection{Blind multichannel deconvolution}

In the blind multichannel deconvolution problems in signal processing,
subspace techniques such as the eigenvector-based algorithm for multichannel blind deconvolution (EVAM) \cite{gurelli1995evam} rely on multichannel annihilation relations. 
Specifically, the multichannel measurement for blind multichannel deconvolution problem is given by
$\widehat f_i =  \widehat s_i \ast \widehat f, \quad i=1,\cdots, N_c$,
where $N_c$ denotes the number of channels, $\widehat s_i$ denotes the unknown convolution kernels, and $\widehat f$ is the unknown underlying signal.
The goal of the blind multichannel deconvolution problem is to estimate $\widehat f$ and $\widehat s_i$ from $\widehat f_i$.

EVAM  \cite{gurelli1995evam}  relies on the cross-channel annihilating filter  relations:
\begin{equation}\label{fourierannihilate}
\widehat{f}_i \ast \widehat{s}_j -  \widehat{f}_j \ast \widehat{s}_i = 0, \quad \forall i\neq j \ .
\end{equation}
Assuming that the convolution kernel $\widehat s_j$ can be represented by
the $d$-tap filter,  the above convolution relations 
can be compactly represented in the matrix form as  
\begin{equation}\label{hankelannihilate}
\begin{bmatrix}
\hank^{\Gamma}_{\Lambda}\left(\widehat f_i\right) \Big| \hank^{\Gamma}_{\Lambda}\left(\widehat f_j\right)
\end{bmatrix}\begin{bmatrix}
\widehat s_j\\-\widehat s_i
\end{bmatrix} =0  \ .
\end{equation}
We now show that the extension of this work to the MRI setting is similar to the popular parallel MRI schemes \cite{shin2014calibrationless,griswold2002generalized}, which were discovered independently without realizing this connection.
\vspace{-1.3em}\subsection{Calibration-based parallel MRI recovery}
\label{multichannel}

The space domain annihilation relations $f_i(\rb) \cdot s_j(\rb) -  f_j(\rb) \cdot s_i(\rb) = 0$ corresponding to \eqref{fourierannihilate} were used in \cite{applause}. The Fourier domain annihilation relations between each pair of channels as in \eqref{hankelannihilate} implies that the matrix formed by linearly stacking the Hankel matrices of $f_i, i=1,\cdots N_c$
\begin{equation}\label{hankmultichannel}
\mathbf L^{\Gamma}_{\Lambda}(f_1,..,f_{N_c}) = \begin{bmatrix}
\hank^{\Gamma}_{\Lambda}\left(\widehat{f_1}\right)| \hank^{\Gamma}_{\Lambda}\left(\widehat{f_2}\right)|\ldots| \hank^{\Gamma}_{\Lambda}\left(\widehat{f_{N_c}}\right)
\end{bmatrix}
\end{equation}
is low-rank{\cite{dologlou1996mri,shin2014calibrationless,Uecker2014espirit,jin2016general,haldar2016p,zhang,mani2017multi}}.

In calibration-based parallel MRI, one often acquires the central Fourier regions in a fully sampled fashion. In this case, one can estimate the null space $\mathbf V$ from the central regions similar to the approach in Fig. \ref{slrmc}.(b). 
Once the null space filters are available, the relation $\mathbf L_{\Gamma}^{\Lambda} \mathbf V=0$ can be used to complete the full matrix \eqref{hankmultichannel} from its partial entries. Consider one column of $\mathbf V$, denoted by $\mathbf v= [\mathbf w_1^T|\mathbf w_2^T|\ldots|\mathbf w_{N_c}^T]^T$; each vector $\mathbf w_i$ is of dimension $|\Lambda|$. The relation $\mathbf L^{\Gamma}_{\Lambda}(f_1,..,f_{N_c}) \mathbf v=0$ can thus be written as 
\begin{equation}\label{onenullspacevector}
\widehat{  f_1}* w_1 + \widehat{  f_2}* w_2 + \ldots \widehat{  f_{N_c}}* w_{N_c} = 0,
\end{equation}
where $*$ denotes linear convolution. 
The above equation implies that the k-space sample of the first coil can be predicted as the linear combination of the nearby samples in all the channels, which is essentially the simultaneous auto calibrating and k‐space estimation (SAKE) model \cite{shin2014calibrationless,griswold2002generalized}. The generalized auto calibrating partially parallel acquisitions GRAPPA approach, which preceded SAKE \cite{shin2014calibrationless,Uecker2014espirit}, can be viewed as an approximation of this linear prediction relation. 
{See \cite{haldar2016p} for a more detailed review of similar multichannel methods.}

\vspace{-1.3em}\subsection{Calibration-less multichannel recovery}
\label{calibrationlesspmri}
The above calibration-based framework is extended to non-uniformly sampled k-space data in \cite{shin2014calibrationless,Uecker2014espirit} using the SLR framework, analogous to the interpolation strategies discussed in Section \ref{interpolation}. The recovery can be interpreted as the rank-deficiency of $\mathbf L^{\Gamma}_{\Lambda}(f_1,..,f_C)$. Accordingly, {\cite{shin2014calibrationless,jin2016general,haldar2016p}} use the  the following low-rank matrix completion to estimate the missing k-space data without any calibration data: 
\begin{eqnarray}\label{eq:SAKE}
\min_{[g_1\cdots g_{N_c}]}  \mathrm{rank}\left(\mathbf L^{\Gamma}_{\Lambda}(g_1,..,g_{N_c})\right) 
~\mbox{subject to }  P_\chi(\widehat g_i) = P_\chi(\widehat  f_i);\quad i=1,\cdots, N_{c}. \notag
\end{eqnarray}
To further exploit the annihilation relationship from the FRI model in addition to the multichannel annihilating relationship, the authors in \cite{jin2016general} solve the optimization
problem \eqref{eq:SAKE} after the k-space weighting.

\vspace{-1em}\section{SLR recovery: fast Algorithms }

We will explain the algorithms in the context of a simple Hankel matrix lifting. {Earlier algorithms  \cite{shin2014calibrationless,Uecker2014espirit,haldar2014low} relied on non-convex rank constrained formulations:
\begin{equation}
\label{eq:loraks}
\underset{\widehat g \in \mathbb{C}^n}{\text{min}} ~ \|P_\chi (\widehat g) - P_{\chi} (\widehat f)\|_2^2  \mbox{  such that  } \rank\left(\hank_{[d]}^{[n]}(\widehat g)\right) \leq r,
\end{equation}
where $r$ is a pre-defined parameter.} Here, $P_{\chi}$ denotes a projection operation to the set $\chi$. Convex unconstrained formulations 
of the form
\begin{equation}
\label{eq:SAKE1}
\underset{\widehat g \in \mathbb{C}^n}{\text{min}} ~\lambda\norm{\hank_{[d]}^{[n]}(\widehat g)}_* + \|P_\chi (\widehat g) - P_{\chi} (\widehat f)\|_2^2.
\end{equation}
where $\|\cdot\|_*$ denotes the matrix nuclear norm and $\lambda$ denotes the regularization parameter are recently being used in applications due to their ability to converge to global minima. 

\vspace{-1.3em}\subsection{Iterative singular value shrinkage}
The formulation in \eqref{eq:SAKE1} can be solved using the singular value thresholding scheme{\cite{qu2015accelerated}}. Specifically, the algorithm alternates between lifting the original signal $\widehat f$ to form $ \Hb(\widehat f)$, followed by singular value {soft} shrinkage of $\Hb(\widehat f)$ to obtain $\mathbf Q$ and un-lifting $\mathbf Q$ to impose the Hankel structure and enforce data consistency. A challenge with this scheme is the high computational complexity and memory demand, especially for multichannel recovery problems.
{In \cite{shin2014calibrationless,haldar2014low}, the authors used the hard thresholding as a singular value shrinkage operation within a similar
iterative optimization framework.}

\vspace{-1.3em}\subsection{UV factorization}
To reduce the computational cost, {the authors in \cite{uvsrebro,jin2016general}} used the following observation:
 \begin{eqnarray}\label{eq:relaxation_nuclear}
 \|\mathbf A\|_* = \min\limits_{\mathbf U,\mathbf V: \mathbf A=\mathbf U\mathbf V^H} \|\mathbf U\|_F^2+ \|\mathbf V\|_F^2,
 \end{eqnarray}
 to realize an SVD-free structured rank minimization algorithm.  
The nuclear norm minimization problem can be solved by replacing the nuclear norm term by the above relation. The constraints are enforced using an alternating direction method of multipliers  (ADMM) algorithm, whose complexity is determined by the rank. So for sparse signals, a significant computation gain can be obtained.

\vspace{-1.em}\subsection{Generic iterative reweighted annihilating filter (GIRAF) algorithm} 
Both of the above strategies rely on the explicit lifting of the signal to the large Hankel matrix, which makes the memory demand and computational complexity of the algorithms high. The GIRAF algorithm avoids the lifting steps altogether and reduces the number of SVD steps; it solves the problem in the original signal domain, thus realizing an algorithm that is comparable in computational complexity to compressed sensing methods \cite{ongie2017fast}. 
{The GIRAF algorithm uses the iterative re-weighted least-squares \cite{irls} strategy, which majorizes the nuclear norm penalty as $\|\mbf A\|^* \leq \left\|\mbf A \mbf Q^{\frac{1}{2}}\right\|_F^2$,
where $\mbf Q = (\mbf A^*\mbf A)^{-\frac{1}{4}}$.
The $i^{\rm th}$ iteration of this algorithm updates $\mathbf Q$ as  $\mathbf Q_{i+1}=\big(\hank_{[d]}^{[n]}(\widehat g)^*\hank_{[d]}^{[n]}(\widehat g_i)\big)^{-\frac{1}{4}}$, followed by the minimization of a quadratic cost function to solve for $g_{i+1}$ obtained by replacing the nuclear norm in \eqref{eq:SAKE1} by $\norm{\hank_{[d]}^{[n]}(\widehat g) \mathbf Q_n}_F^2$.
The GIRAF algorithm relies on fast Fourier transforms to evaluate the matrix vector product between $\hank_{[d]}^{[n]}(\widehat g) $ and each column of $\mathbf Q$; this approach exploits the convolutional structure of $\hank_{[d]}^{[n]}(\widehat g) $. We note that this approach does not require the creation or storage of the large structured matrix, but works directly with the signal samples of $g$. The MATLAB implementation of the GIRAF algorithm is available at https://github.com/cbig-iowa/giraf. }

\vspace{-1.3em}\subsection{Algorithms for non-convex SLR formulations}
All of the above algorithms are designed for convex formulations \eqref{eq:SAKE1}. The early algorithms \cite{shin2014calibrationless,Uecker2014espirit} relied on successively solving 
\begin{equation}
\label{eq:girafloraks}
g_{n+1} = \underset{\widehat g \in \mathbb{C}^n}{\text{min}} ~ \|P_\chi (\widehat g) - P_{\chi} (\widehat f)\|_2^2  + \lambda \|\widehat g - \mathcal M_r(\widehat g_n) \|^2,
\end{equation}
where $M_r(\widehat g_n)$ is the approximation of the $n^{\rm th}$ iterate $\widehat g_n$ such that $\hank_{[d]}^{[n]}(\widehat g)$ is of rank $r$. Recently, the multiplicative majorization strategy in the GIRAF algorithm \cite{ongie2017fast} has been adapted into the LORAKS setting to realize faster algorithms.

\vspace{-1em}\section{Generalization to Machine Learning}
\subsection{Recovery of point clouds on surfaces: non-linear generalization of union of subspaces model}
\label{pointclouds}
We will now consider a non-linear generalization of the FRI theory discussed above, which will facilitate the recovery of surfaces or points living on surfaces from a few noisy samples. The main motivation is the joint recovery of an ensemble of images (e.g.g images in a cardiac MRI time series) from their noisy and undersampled measurements. This approach is a non-linear generalization of the popular union of subspaces model, which represents the images as a sparse linear combination of some basis images.

\begin{figure}[t!]
	\center
	\subfigure[Illustration]{\includegraphics[width =0.55\textwidth]{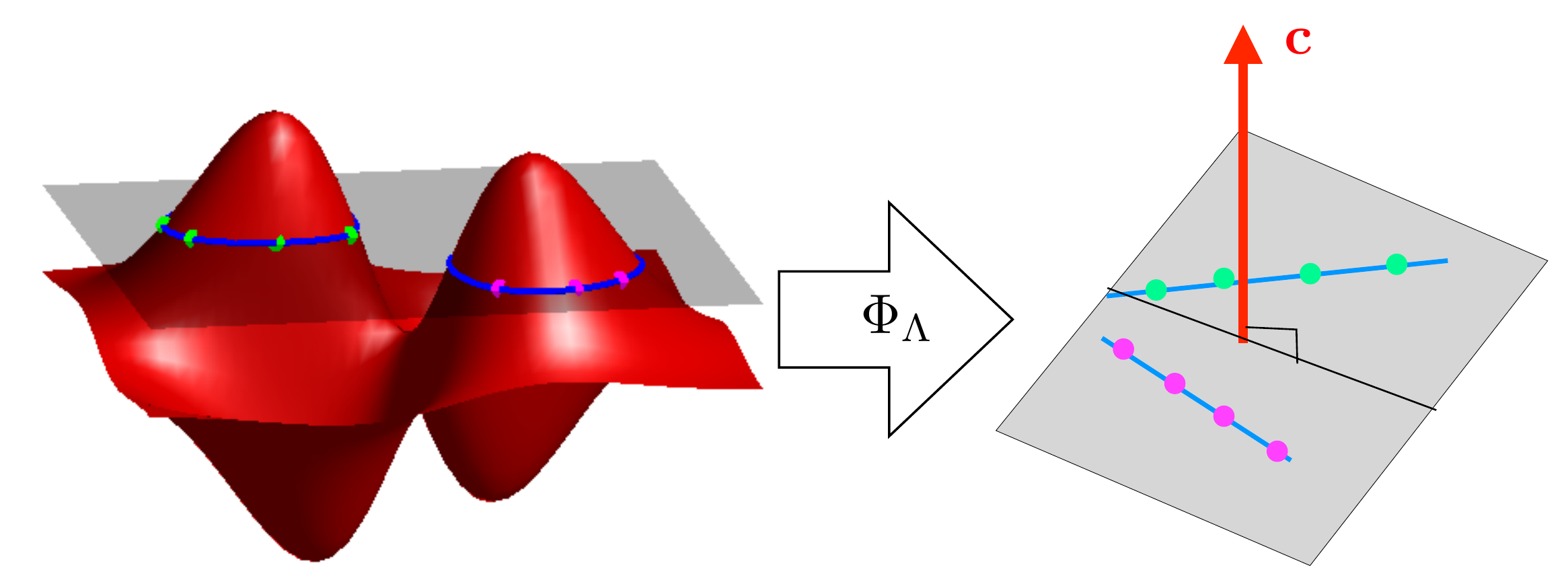}}\hspace{1em}
	\subfigure[Denoising]{\includegraphics[width =0.4\textwidth]{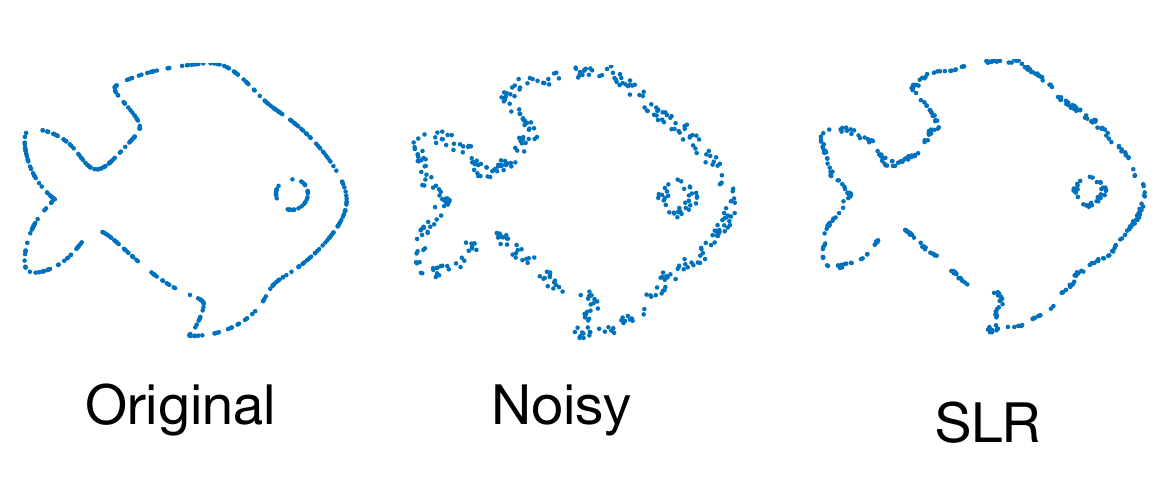}}
	\caption{Illustration of the non-linear lifting operation in 2-D, which will map closed union of bandlimited curves to union of sub-spaces. (a) Each closed curve is mapped to a subspace. The SLR scheme relies on the low-rank structure of the subspace where the lifted points live in to recover the curves or denoise the points living on union of bandlimited curves. The utility of this scheme in the denoising of shapes is illustrated in (b). The recovery is posed as a nuclear norm minimization of the feature maps, which was solved using an iterative reweighted least-squares algorithm exploiting the kernel trick \cite{manifold1}. This approach is extended to higher dimensions to recover free breathing and ungated cardiac MRI data, illustrated in Fig. \ref{DMRI}. }
	\label{FigNonLinear}
\end{figure}

We model the surface as the zero level set of a bandlimited function as in Section \ref{2dtheory}. 
For simplicity, we will explain the approach in 2D. Consider an arbitrary point $\mathbf x$ on the curve  $\widehat h(\mathbf x)=0$, where $h(\mathbf x)$ is specified by \eqref{bl}. The space domain annihilation relation translates to $\mathbf c^T \phi_{\Lambda}(\mathbf x)=0$, where
\begin{equation}\label{featurematrix}
\phi_{\Lambda}(\mathbf x)=
 \begin{bmatrix} \exp\left({j\,\mathbf k_1^T}\mathbf x\right)&
\ldots&
\exp\left(j\,\mathbf k_{|\Lambda|}^T\mathbf x\right)
\end{bmatrix}^T 
\end{equation}
is a non-linear mapping or lifting of a point $\mathbf x$ to a high-dimensional space \cite{manifold1} (see Fig. \ref{FigNonLinear}) of dimension $|\Lambda|$. Since this non-linear lifting strategy is similar to feature maps used in kernel methods \cite{kernelbook}, we term  $\phi_{\Lambda}(\mathbf x)$ as the feature map of the point $\mathbf x$. 

Let us now consider a set of $N$ points on the curve, denoted by $\mathbf x_1,\cdots,\mathbf x_N$. The above annihilation relations can be compactly represented as $\mathbf c^T \Phi_{\Lambda}(\mathbf X) = 0$, where
\begin{equation}
\label{annnl}
\Phi_{\Lambda}(\mathbf X) = \begin{bmatrix}
\phi_{\Lambda}(\mathbf x_1) & \phi_{\Lambda}(\mathbf x_2) & \ldots &\phi_{\Lambda}(\mathbf x_N) 
\end{bmatrix} 
\end{equation}
is the feature matrix of the points.  The results in \cite{manifold1} show that if the number of samples exceeds a bound that depends on the complexity of the curve, it can be uniquely estimated. 

We have shown in \cite{manifold1} that if we choose a feature map with a bandwidth of $\Lambda_1$ such that $\Lambda \subset \Lambda_1$, the feature matrix satisfies the relation similar to \eqref{rankbound}:
\begin{equation}
{\rm rank}\left(\Phi_{\Lambda_1}(\mathbf X)\right) \leq ~~ |\Lambda_1|-|\Lambda_1\ominus \Lambda|.
\end{equation}
Generalizing the SLR approach in Section \ref{interpolation}, we recover the samples on the curve/surface from a few measurements corrupted by noise \cite{manifold1} by solving 
{an optimization problem:
\begin{equation}
\label{klr}
\underset{\mathbf X}{\text{minimize}} ~\lambda\norm{\Phi_{\Lambda_1}(\mathbf X)}_*  + \|\mathcal A(\mathbf X)-\mathbf b\|^2.
\end{equation}
We use an iterative-reweighted least-squares (IRLS) algorithm \cite{irls} that relies on the \emph{kernel trick} \cite{kernelbook} for recovery. Specifically, the algorithm alternates between a quadratic optimization scheme and the evaluation of the graph Laplacian matrix; the algorithm may be interpreted as a the discretization of the manifold by a graph, whose connectivity is specified by the distances on the manifold/surface. Note that unlike the approaches in Section \ref{interpolation}, this algorithm is non-convex. {Hence, the above optimization problem cannot be guaranteed to achieve the global minimum.} Despite the lack of guarantees, the algorithm was able to yield good performance in applications, as seen in Fig. \ref{FigNonLinear} and Fig. \ref{DMRI}.}

\vspace{-1.3em}\subsection{Relation to Deep Convolutional Neural Networks}

One of the interesting twists of the SLR approach is its relation to deep neural networks.
{Specifically, in the recent  theory of  deep convolutional framelets \cite{ye2018deep}, the authors showed that a deep neural network can be interpreted as a framelet representation
whose basis can be obtained from  Hankel matrix decomposition.
Moreover, in a recent follow-up study \cite{ye2019understanding}, the authors further revealed
that the recified linear unite (ReLU) plays a key role in making the representation adaptive to input {by providing
combinatorial basis selection}.}

Specifically, for a given Hankel matrix $\Hb(f)$,
{let us consider two  matrices  $\Phib^\top$  and $ \Psib$ whose dimensions are determined such that
they can  multiplied to the left and the right
of the Hankel matrix, respectively.
Furthermore, suppose there exists matrices
 $ \tilde\Phib$ and their duals $\tilde \Psib$ 
satisfying the frame condition}
 $\tilde \Phib \Phib^\top = \Ib,\quad \Psib \tilde \Psib^{\top} =\Ib, $
 where  the superscript $^\top$ denotes the matrix transpose and $\Ib$ refers to an identity matrix with appropriate size.
 {The existence of such matrices can be trivially shown as infinite number of orthonormal matrices satisfy the frame condition.
 In fact, the frame condition allows overcomplete representation.} 
 {Then, it is easy to show 
\begin{eqnarray}
\hank(f) = \tilde\Phib \Cb \tilde \Psib^\top&,& 
\mbox{where}\quad \Cb =  \Phib^\top \hank(f) \Psib .  \label{eq:enc}
\end{eqnarray}
}
One of the most interesting observations in \cite{ye2018deep} is that \eqref{eq:enc}  can be equivalently represented by encoder
and decoder convolution structure:
\begin{eqnarray}
\Cb =  \Phib^\top \left( f \ast  \alpha(\Psib)\right) &, &
f  = \left(\tilde\Phib \Cb\right) \ast \beta(\tilde \Psib)   \label{eq:pr}
\end{eqnarray}
where $\alpha(\Psib)$ and $\beta(\tilde\Psib)$ are
the encoder and decoder layer multichannel convolution filters obtained by rearranging $\Psib$ and $\tilde\Psib$, respectively \cite{ye2018deep}. 
This observation led to the findings  that $\Phib^\top$ and $\tilde \Phib$ correspond to the pooling and unpooling layers, respectively \cite{ye2018deep}.

{
However, to satisfy the frame conditions, 
the number of output filter channels  should increase
exponentially, which is difficult to satisfy in practice  \cite{ye2018deep}. 
Moreover, in contrast to SLR approaches, the exact decomposition of the Hankel matrix in \eqref{eq:pr}
 is not interesting in neural networks, since the output of the network 
should be different from the input due to the task-dependent processing.
Moreover,  the  representation should be generalized well for various inputs rather than a specific input at the training phase.
In a recent extension of the deep convolutional framelets \cite{ye2018deep}, the authors revealed that the convolutional framelet representation is indeed combinatorial 
 due to the combinatorial nature of ReLU. Specially, for the case of an encoder-decoder convolutional neural network (CNN) without a skipped connection
 it was shown that the CNN ouput $g$ can be represented
as follows \cite{ye2019understanding}:
\begin{eqnarray}\label{eq:basis}
g 
&=& \sum_{i} \langle {\bb}_i(f), f \rangle \tilde  {\bb}_i(f),
\end{eqnarray}
where $ {\bb}_i(f)$ and $\tilde  {\bb}_i(f)$
denote the $i$th columns of the $\Bb(f)$ and $\tilde\Bb(f)$ matrices given by
\begin{eqnarray}
\Bb(f)&=& \Eb^1\Sigmab^1(f)\Eb^2 \cdots  \Sigmab^{\kappa-1}(f)\Eb^{\kappa},~\quad \label{eq:Bc}\\
\tilde \Bb(f) &=& \Db^1\tilde\Sigmab^1(f)\Db^2 \cdots  \tilde\Sigmab^{\kappa-1}(f)\Db^{\kappa}, \label{eq:tBc}
\end{eqnarray}
where $\Sigmab^l(f)$ and $\tilde\Sigmab^l(f)$ denote diagonal matrices with 0 and 1 values that are determined by the ReLU output
in the previous convolution steps;
$\Eb^l$ (resp. $\Db^l$) denotes the $l$-th layer encoder (resp. decoder) matrix that is composed of pooling matrix $\Phib^l$ (unpooling matrix
 $\tilde\Phib^l$) and encoder filter matrix $\Psib^l$ (resp. decoder filter matrix $\tilde\Psib^l$).
Similar basis representation holds for the encoder-decoder CNNs with skipped connection~\cite{ye2019understanding}.
}

{When there are no ReLU nonlinearities and pooling and filter matrices satisfy the frame condition
 for each $l$,
the representation \eqref{eq:basis} is indeed a convolutional frame representation of $f$
as in \eqref{eq:pr},
ensuring perfect signal reconstruction.
{Even with the ReLU, there exists conjugate filter sets that can allows perfect reconstruction condition
\cite{ye2018deep}.}
However, 
the explicit  dependence  of \eqref{eq:Bc} and \eqref{eq:tBc} on the input $f$
due to the ReLU nonlinearity {makes the neural network representation much richer and  adaptive to different input signals, since}
the resulting frame representation in \eqref{eq:basis} depends on specific input. 
{In this regard, the role of ReLU may be similar to the sparsity patterns in compressed sensing MRI, since the sparsity pattern
determines the difference basis representation of CS for given inputs.}
Furthermore, 
the number of distinct linear representations
 increases exponentially
with the network depth, width, and the skipped connection,  thanks to the combinatorial
nature of ReLU nonlinearities~\cite{ye2019understanding}. 
This exponentially large  {\em expressivity} of the neural network 
is another important advantage, which may, with the combination of the aforementioned {\em adaptivity},  explain the origin
of the success of deep neural networks for image reconstruction.}


\vspace{-1em}\section{Application of SLR for improved MRI}
The main benefit of the SLR framework is its ability to account for a wide variety of signal models, facilitating the exploitation of extensive redundancies between their Fourier samples. This feature makes SLR applicable to a wide variety of MRI applications, listed below.

\vspace{-1.3em}\subsection{Highly accelerated MRI}
During the past few years, different SLR priors were introduced for highly accelerated MRI, each designed to exploit specific signal properties. 
\subsubsection{{Support/sparsity priors \cite{haldar2014low}}}
\label{loraks}
The signal is assumed to be sparse or support limited to a region in low-rank modeling of local k-space neighborhoods (LORAKS), which results in annihilation conditions in k-space. The LORAKS  scheme formulated the CS-MRI problem by using the block Hankel matrix of the images as the prior. The LORAKS framework also exploits phase constraints, which are detailed in \cite{haldar2014low}. 

\subsubsection{{Transform domain sparsity {\cite{jin2016general,ongie2016off,liangslr}}}}
The annihilating filter-based low-rank Hankel matrix (ALOHA) approach  considers general signals  that can also be  sparse in a transform domain. Specifically, the signal $f$ is modeled as $
\mathrm{L} f = w$, where $\mathrm{L}$ denotes a constant coefficient linear differential equation, often called the whitening operator \cite{ye2016compressive}: 
\begin{equation}\label{eq:L0}
\mathrm L := b_K\mathrm \partial^K+b_{K-1}\mathrm \partial^{K-1}+\ldots+b_1\mathrm \partial+b_0 , 
\end{equation}
and 
$w$ is an innovation signal composed of a stream of Diracs or differentiated Diracs. Evaluating the Fourier transform, we have $\widehat w: = \widehat l \widehat f $
where 
$	\widehat l(f) =  b_K (i2\pi f)^K +\ldots+b_1(i2\pi f)+b_0$ 
and the associated Hankel matrix  $\hank(\widehat w)$  from the innovation spectrum $\widehat w$ 
becomes rank deficient. The Fourier coefficients are interpolated similar to \eqref{eq:nucmin} by minimizing
the nuclear norm of a block Hankel matrix whose entries are the Fourier coefficients of $\hat f$ weighted by $\hat l$. 

\subsubsection{{Piecewise smooth signal model {\cite{ongie2017convex,liangslr}}}}
\label{smooth} The GIRAF algorithm generalized the {piecewise polynomial} 1-D model \cite{liangslr} to recover piecewise constant multi-dimensional signals from their sparse Fourier coefficients by minimizing the nuclear norm of the vertically stacked block Hankel matrices in \eqref{lifted} (see Figure \ref{Figstacking}.(a)),  as described in Section \ref{2dtheory}. Each of the block Hankel matrices correspond to weighted Fourier coefficients of the signal. 

\begin{figure}[t!]
	\centering
\includegraphics[width =0.7\textwidth]{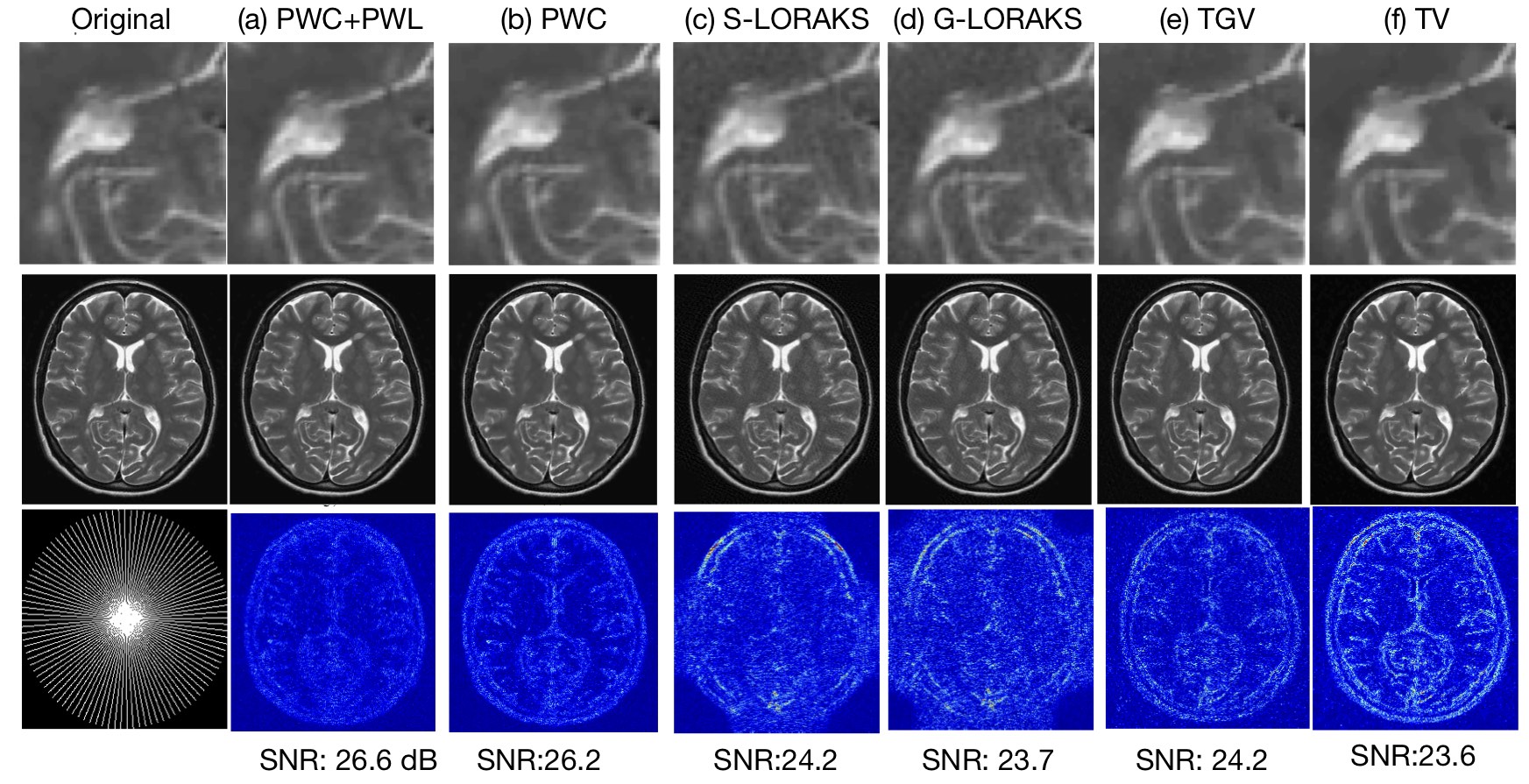}~~\vspace{-1.1em}
	\caption{Illustration of different flavors of SLR recovery in a single coil setting. The Fourier transform of the image was sampled on the radial grid, which corresponds to an acceleration of 4.85. The columns correspond to (a) GIRAF, which assumes a piecewise constant image model, where the nuclear norm of the block Hankel matrix in \eqref{lifted} is minimized, (b) GSLR, where the image is modeled as the sum of a piecewise constant and piecewise linear functions as in \eqref{gslreq}, (c) S-LORAKS and  (d) G-LORAKS \cite{haldar2014low}, (e) total generalized variation (TGV), which is a discrete model that represents the images as the sum of piecewise linear and piecewise constant factors, and (f) total variation (TV). We note that the Fourier domain models (a)\&(b) offer better reconstructions compared to their discrete counterparts (f) and (e), respectively, which can be appreciated from the zoomed images as well as the error images. We note that  both (a)  and (b) consider the low-rank structure of Hankel matrices constructed from weighted Fourier samples; this allows these methods to exploit the continuous sparsity of the edges, similar to discrete approaches such as TV or wavelet methods. The performance of the algorithms can be further improved by combining multiple SLR priors to exploit different signal priors (e.g., piecewise smoothness, multichannel sampling, smoothness of phase) \cite{mani2017multi}. These combinations can be either accounted for by different regularization terms or combined into a single nuclear norm penalty of a more complex structured matrix obtained by vertical and horizontal stackings to exploit the redundancies. }\vspace{-2.1em}
		\label{gslr}
\end{figure}

This model was recently extended to represent the image as a linear combination of piecewise constant ($f_{\rm pwc}$) and piecewise linear components ($f_{\rm pwl}$)\cite{hu2018generalized}: $
f(\mathbf x) = f_{\rm pwc}(\mathbf x) + f_{\rm pwl}(\mathbf x)$. 
The recovery is posed as:
\begin{eqnarray}\nonumber
\min_{g_1,g_2\in \Cd^{n} } & \| \mathcal T (\widehat {\nabla g_1})\|_*  + \| \mathcal S (\widehat {\nabla_2 g_2})\|_* + \lambda \|P_\Omega(\widehat{g_1}+\widehat{g_2}) - P_\Omega(\widehat f)\|^2
\label{gslreq},
\end{eqnarray}
where $\mathcal S$ is the matrix obtained by the vertical stacking of the block Hankel matrices of three second degree partial derivatives, denoted by the vector $\widehat{\nabla_2 f}$. We demonstrate the benefit of some of the flavors of the proposed scheme in Fig. \ref{gslr}

\vspace{-1.3em}\subsection{{Calibration-free correction of trajectory and phase errors}}  

The multichannel framework introduced in Section \ref{calibrationlesspmri} provides a versatile tool to solve several calibration problems in MRI, where different k-space regions are acquired from different excitations; these datasets will differ in phase errors, which often manifest as artifacts. As discussed in Section \ref{multichannelmodel}, \eqref{eq:parallel} can also be used to model the acquisition of multi-shot Fourier data. Specifically, the data is split into multiple groups or virtual channels, each corresponding to different distortions. The joint recovery of these groups or virtual channels are performed as in \eqref{eq:SAKE}.



\subsubsection{{Correction of phase errors in multi-shot diffusion MRI {\cite{mani2017multi}}}}
\label{diffusionmri}
\begin{figure}[h!]\vspace{-1em}
	\centering
	\subfigure[Artifact correction in diffusion MRI]{\includegraphics[trim = 0mm 30mm 0mm 0mm, clip,height =0.35\textwidth]{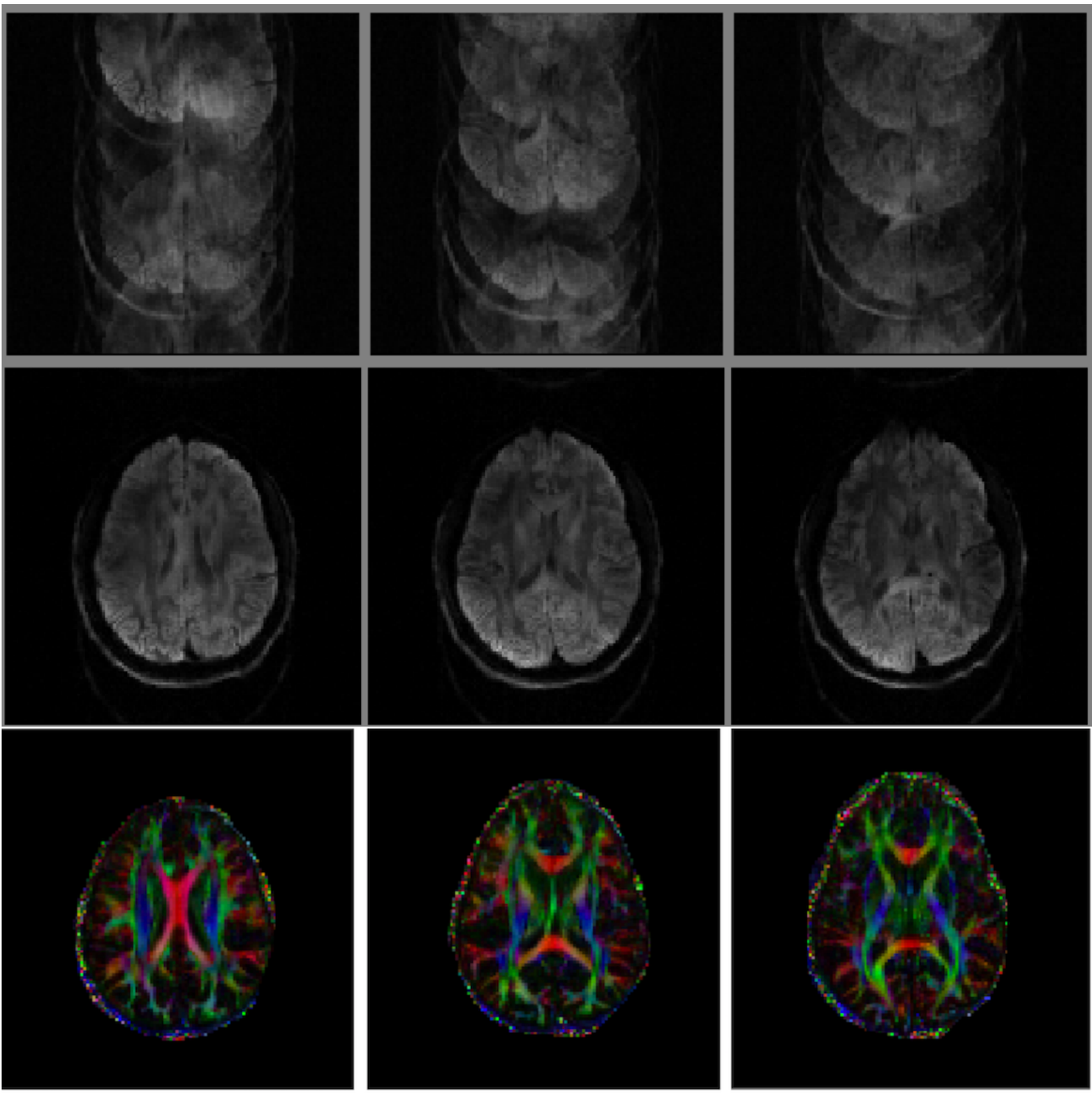}}~~
	\subfigure[B0 distortion correction in EPI]{\includegraphics[height =0.35\textwidth]{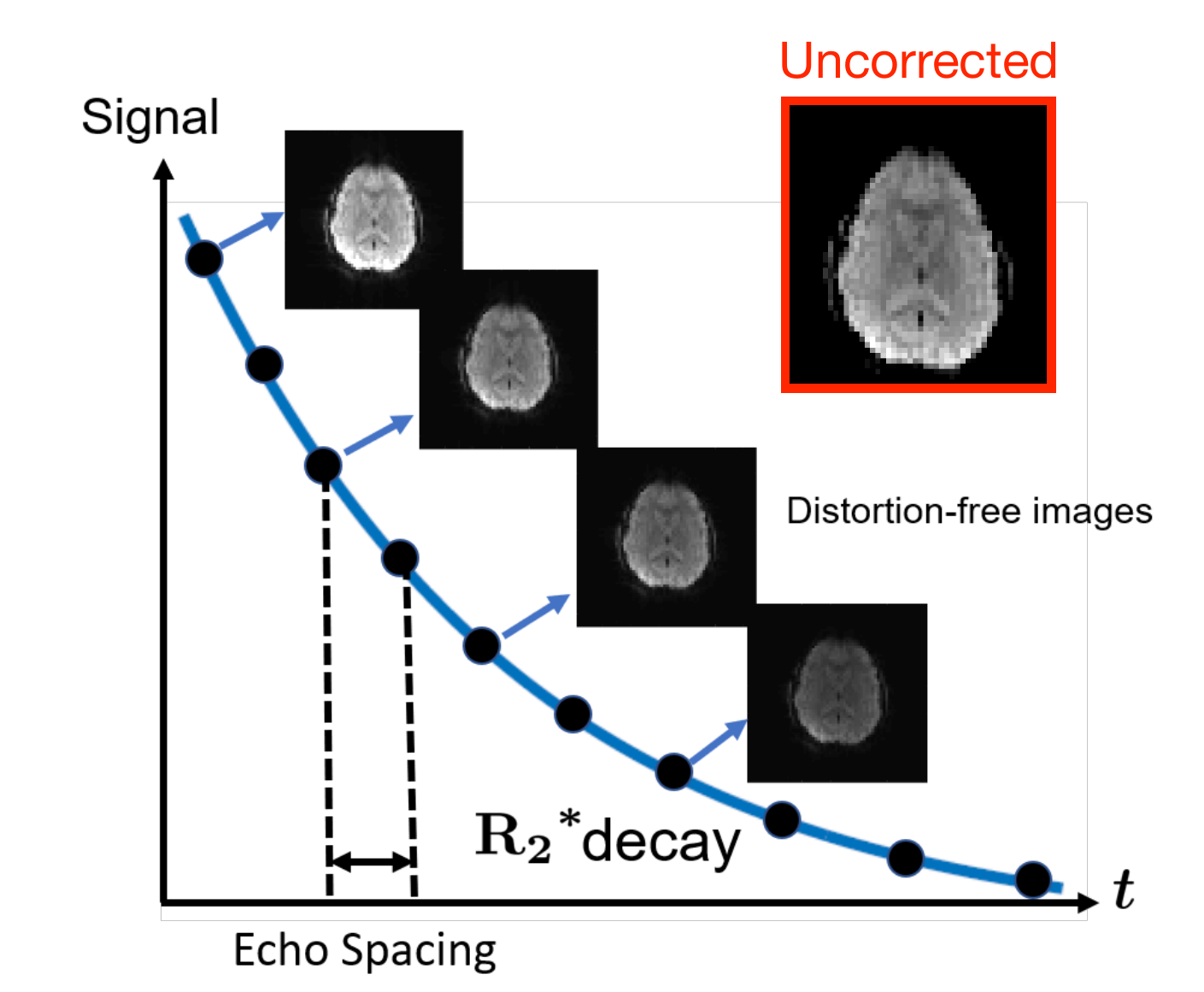}}~~
	\caption{Example applications enabled by the SLR framework. (a) High-resolution four-shot diffusion MRI enabled by the SLR framework. The top row corresponds to uncorrected four-shot diffusion weighted MRI data from a single diffusion direction. Subtle subject motion between the shots results in phase errors between the measured Fourier datasets. In addition, eddy current artifacts also manifest as shifts in Fourier space between odd/even lines. These errors manifest as Nyquist ghosting artifacts in the uncorrected  diffusion wighted images (DWI) in the top row. The SLR algorithm called MUSSELS \cite{mani2017multi} recovers the eight images, corresponding to each shot and odd/even lines separately, exploiting the phase relations between the images. The sum-of-squares combined DWI data from the above eight images for each slice are shown in the second row. The information from sixty such directions provides the fractional anisotropy diffusion MRI maps shown in the bottom row. (b) Correction of B0 distortions in EPI data \cite{arvindB0}: the long readouts in EPI often result in spatial distortions, resulting from the inhomogeneity of the main (B0) magnetic field. Ignoring the B0-induced magnetization evolution during the long EPI readouts will be associated with spatial distortions. We reformulated the EPI distortion correction as a time-series recovery problem, where multiple images corresponding to different segments of the readouts are recovered exploiting the exponential structure of the signal. Note that each of the segments are highly undersampled. The missing Fourier samples are filled in using a structured low-rank matrix completion. }
	\label{diffusion}
\end{figure}

Diffusion MRI (DMRI) is a valuable tool for assessing brain connectivity and tissue micro-structure. High-resolution DMRI is often acquired using multi-shot EPI schemes, where different Fourier regions are acquired from different radio-frequency excitation pulses. For example, the even lines in a two-shot EPI sequence are collected from one shot, while the odd lines are collected in the second shot. Subtle physiological motion between the shots in the presence of diffusion encoding gradients manifest as motion induced phase errors between the shots. Previous works  \cite{mani2017multi} have shown that these phase errors can be corrected using the SLR scheme, even when the data is acquired using 4-8 shots. The results in Fig. \ref{diffusion}.(a) shows the improved reconstructions offered by this scheme.

\subsubsection{Compensation of trajectory errors \cite{jiangmrm}}
	MRI images can also suffer from artifacts, resulting from k-space data of different excitations experiencing different distortions. A typical example is radial imaging, where the shifted radial spokes cause streaking artifacts. A matrix completion using the low-rank prior \cite{jiangmrm} is used to jointly recover the artifact-free images and the uncorrupted calibration data. Several other groups have shown the utility of the low-rank based approach for the correction of trajectory errors in EPI and radial setting (not referenced here due to space constraints).


\vspace{-1.3em}\subsection{{Correction of k-space outliers \cite{bydder,jin2017mri}}}
Many MRI artifacts from the instability of an MR system, patient motion, inhomogeneities of gradient fields, etc. are manifested as outliers in k-space data. Because MR artifacts usually appear as sparse k-space components, the artifact-corrupted MRI measurements $\widehat z(f)$ can be modeled as $
\widehat{z}(f) = \widehat{x}(f) + \widehat{s}(f)$, where $\widehat{x}(f)$ is a k-space data of an artifact-free image and $\widehat{s}(f)$ is  the sparse k-space outlier \cite{jin2017mri}. If the unknown signal can be sparsified by applying a whitening operator \eqref{eq:L0} by performing a lifting to a Hankel structured matrix, we can see that
\begin{eqnarray}\label{eq:hankel_rpca}
\hank(\widehat y) = \underbrace{\hank(\widehat l \odot \widehat x)}_{\mbox{low-rank}}+\underbrace{\hank(\widehat l \odot \widehat s)}_{\mbox{sparse}}, 
\end{eqnarray} 
where $\widehat l$ denotes the spectrum of the whitening operator.

\begin{figure}[t!]
	\centering	
	\subfigure[Sparse + low rank decomposition.]{\includegraphics[width =0.24\textwidth]{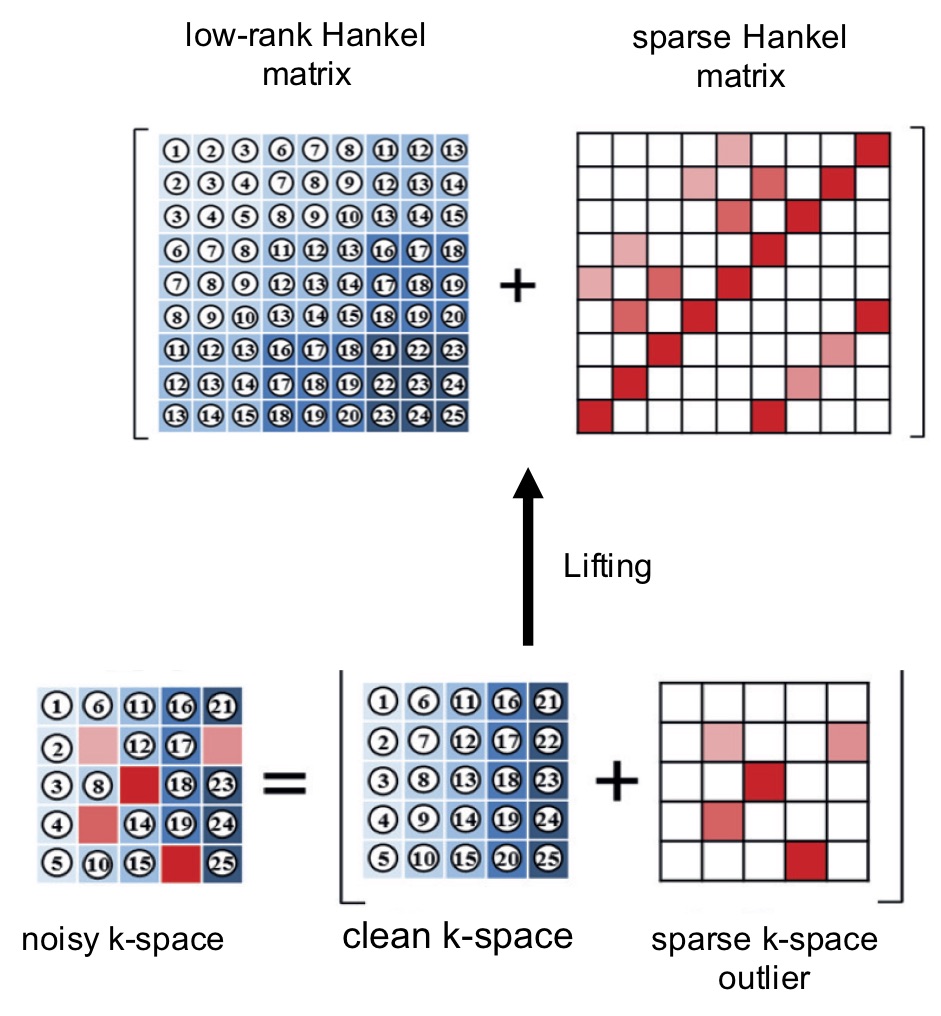}} \hspace{1cm}
	\subfigure[MR k-space artifact removal example.]{\includegraphics[width =0.4\textwidth]{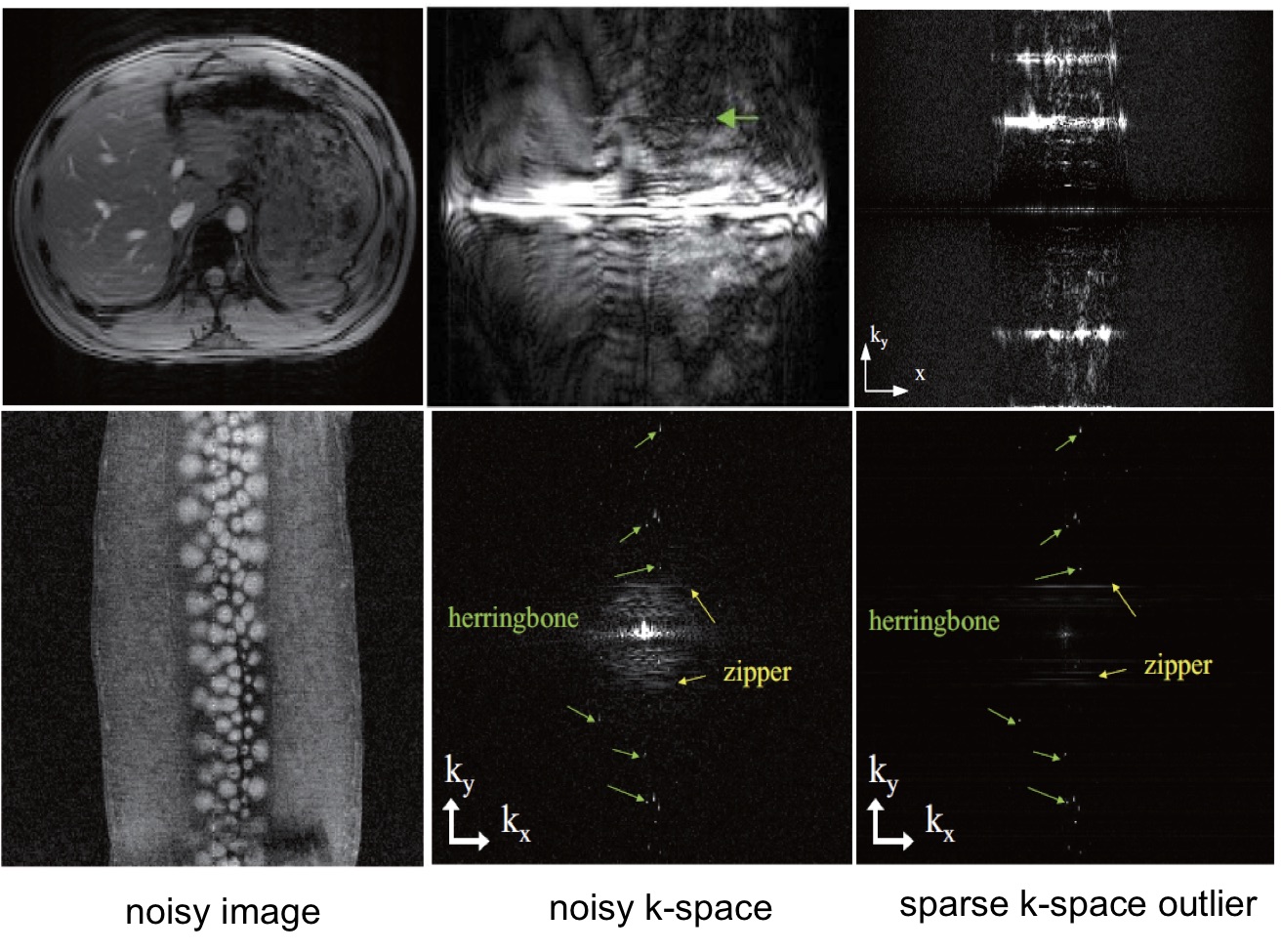}}~~
	\caption{(a) Lifting of image with sparse outlier. Noiseless image becomes low-rank Hankel matrix, whereas the noisy image becomes a sparse Hankel matrix so that robust principal component analysis can decompose the sparse and low-rank Hankel matrices. Then, by unlifting the Hankel matrix, we can obtain the noiseless k-space data and the clean images. (b) Example  of MR artifact removals: (left) noisy images, (center) noisy k-space, and (right) decomposed sparse k-space outliers.}\vspace{-2em}
	\label{sparselift}
\end{figure}

Note that the second term in Eq. \eqref{eq:hankel_rpca} is sparse, because
the lifted Hankel matrix from sparse components is still sparse, as illustrated in Fig.~\ref{sparselift}.
Thus, Eq. \eqref{eq:hankel_rpca} becomes a structure for sparse + low-rank decomposition, and
a modified version of  robust principal component analysis (RPCA) was
used to decompose the low-rank and sparse component of the Hankel matrix  $\hank(\widehat y)$.
After the sparse + low-rank decomposition, the weighted k-space for the low-rank component is returned to original k-space by performing an un-weighting \cite{jin2017mri}.


\vspace{-1.3em}\subsection{Recovery of exponentials}

\subsubsection{{Parameter mapping \cite{morasa,bilgic}}} MR parameter mapping methods, which estimate the $T_1$ and $T_2$ relaxation constants, can enable tissue characterization and hence are clinically very valuable. However, the the long scan time resulting from the need for multiple temporal frames makes these schemes challenging for routine clinical use. Under-sampling each temporal frame, followed by sparse recovery, has been a popular approach to reduce scan time. The spatio-temporal signal can be modeled as an exponential with smoothly varying parameters as described in Section \ref{expslr}, which facilitates its recovery from under-sampled measurements. 

\subsubsection{{Field inhomogeneity compensation \cite{nguyen,arvindB0}}}
\label{b0section}
MRI schemes such EPI are associated with long readouts. The EPI signal at the time point $t$ after the excitation is the Fourier sample of the image modulated by a time-evolving exponential: $
\rho[\mathbf{r},n] = f(\mathbf{r})\; \beta(\mathbf{r}) ^{n}$, where the exponential parameter $\beta(\mathbf r)= e^{-(\mathbf{R}_{2}^{*}(\mathbf{r})+j \omega(\mathbf{r}))T}$ results from the field inhomogeneity and $T$ is the sampling rate in time; $f(\mathbf r)$ is the true image. The standard inverse Fourier transform reconstruction ignoring the exponential term thus recovers the field inhomogeneity distorted image. We used the approach described in Section \ref{expslr} to recover the entire time series $\rho[\mathbf{r},n] $ from the Fourier measurements. Post-recovery, the image $\rho[\mathbf{r},0] $ is chosen as the un-distorted signal. Since there are two complex unknowns $f(\mathbf r)$ and $\beta(\mathbf r)$ at each pixel, we use two shifted EPI readouts to recover the time series \cite{arvindB0}. The results of this approach are shown in Fig. \ref{diffusion}.(b).

\subsection{Free-breathing and ungated cardiac MRI  \cite{manifold1}}
\label{storm}
In cardiac MRI clinical practice, breath-held cardiac cine MRI is the standard protocol to evaluate cardiac function. Many subject groups cannot tolerate the breath-holds, which disqualifies such patients for cardiac MRI exams. While free-breathing and ungated cardiac MRI is the ideal protocol, a main challenge is the quite significant acceleration needed to facilitate this scheme. The work in \cite{manifold1} used the surface recovery strategy in Section \ref{pointclouds} to recover free-breathing and ungated data from highly undersampled measurements with great success as shown in Fig. \ref{diffusion}.(b). 

\begin{figure}[t!]
	\centering
	\subfigure[Comparison with XD-GRASP]{\includegraphics[width =0.41\textwidth]{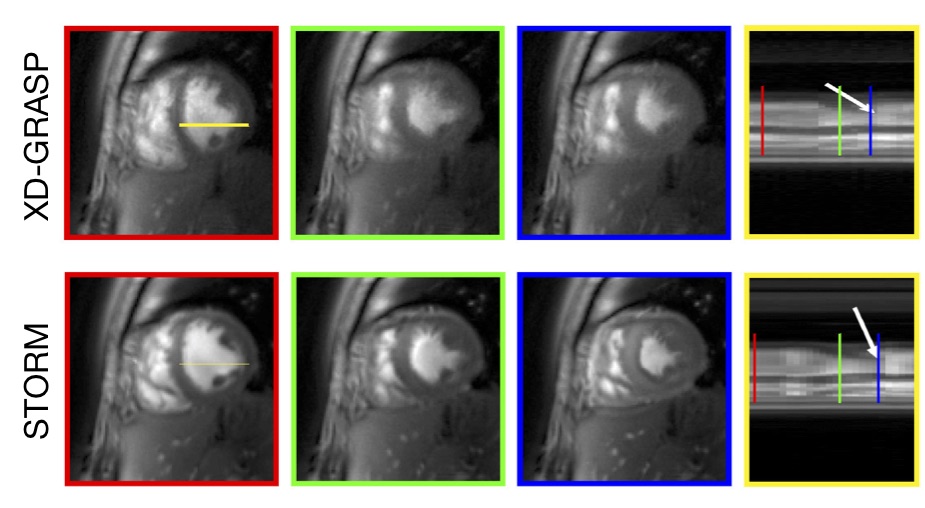}}
	\subfigure[Comparison with breath-held cine]{\includegraphics[width =0.37\textwidth]{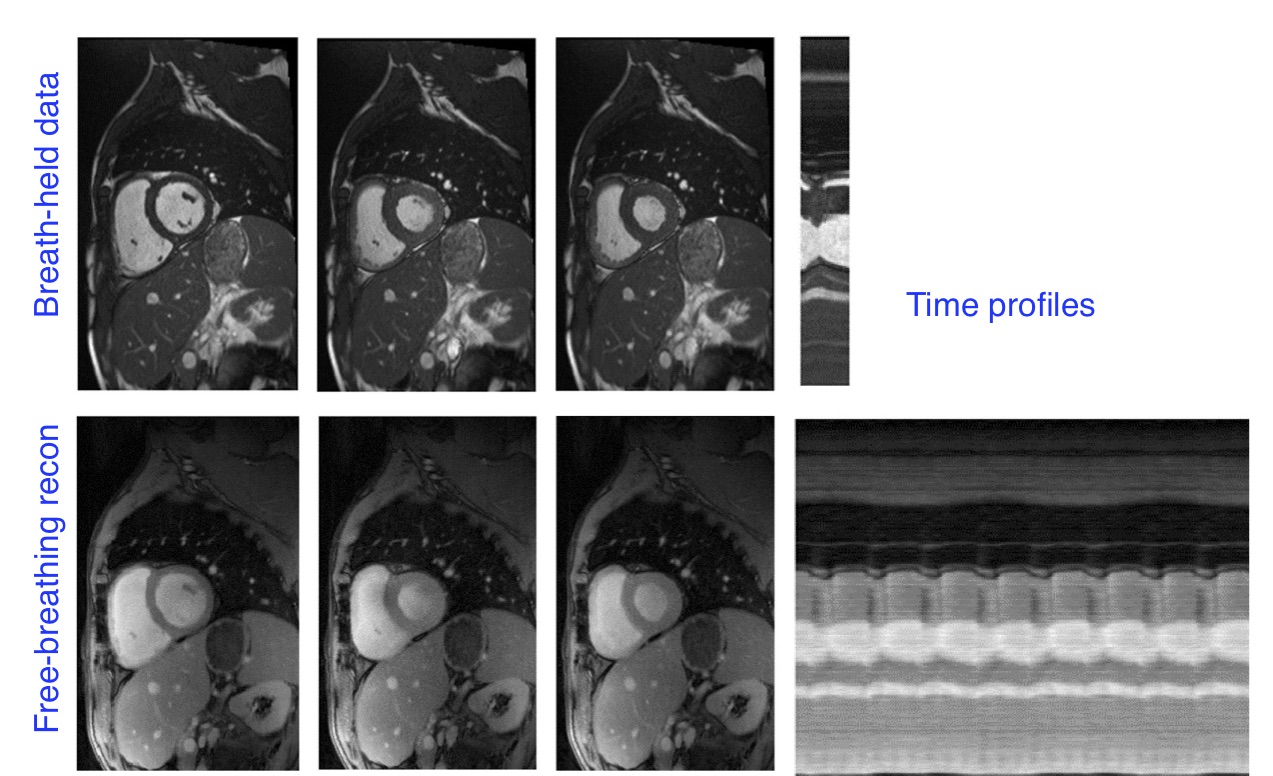}}~~
	\caption{Illustration of SToRM algorithm \cite{manifold1} described in Section \ref{storm}. The SToRM algorithm exploits the manifold structure of free-breathing and ungated images in high-dimensional space. Specifically, each image can be viewed as a smooth function of two parameters: the cardiac phase and the respiratory phase. A non-linear transformation specified by \eqref{annnl} transforms the data to a subspace. We recover the images from their undersampled measurements by exploiting  this structure, which is compactly captured by the nuclear norm of the feature matrix. We solve the optimization problem using an IRLS algorithm that uses the kernel trick, which eliminates the need to explicitly evaluate the complex features. (a) shows the comparisons of SToRM reconstructed images with XD-GRASP, which is an explicit binning strategy; XD-GRASP bin the images to distinct cardiac/respiratory phases, followed by the total variation recovery of the images. We observe that the implicit binning of the data offered by SToRM results in reduced blurring and improved fidelity (b) shows the comparisons of SToRM reconstructions (bottom row) with classical breath-held acquisitions (top-row) that bins the data from different cardiac cycles. We observe that the image quality is comparable, whole SToRM offers real-time imaging capabilities, allowing us to visualize cardiac and respiratory functions simultaneously.
		(a) Comparison}
	\label{DMRI}\vspace{-2em}
\end{figure}

\vspace{-1em}\section{Conclusion \& {Future Work}}
The SLR formulation provides a flexible framework to exploit different continuous domain signal priors, which are difficult for current discrete compressive sensing frameworks to capture. The framework comes with theoretical guarantees and fast algorithms and software, making it readily applicable in multidimensional imaging problems including MRI. The recent results showed that there are important links between SLR frameworks and machine learning approaches. The proposed framework is demonstrated in several challenging MRI applications.
\label{sec:conclusion}

{In spite of its flexibility and many applications, there are still remaining open challenges, some of which
will be discussed below.\\
\textbf{Sampling patterns:} The single-channel theoretical results \eqref{eq:samp_comp} and \eqref{probrec} guarantee the recovery of the Fourier coefficients of the signals on a grid from a randomly chosen subset of coefficients. The results exhibit a $\log^4$ dependence on the final grid size, suggesting increasing sample demand with larger grid sizes. However, we note from Sections \ref{1dfri} and \ref{pwcsection} that once the null space is identified, the signal extrapolation to any grid size is possible from the low-frequency Fourier samples \cite[Theorem 1]{ongie2016off}. This strongly suggests a variable-density sampling approach, which may significantly reduce the sampling demand,  and which was also confirmed by the empirical results in \cite{ongie2017convex}. More theoretical work in this area is needed to determine the best sampling patterns. \\
While empirical results demonstrate the great benefit of SLR schemes in multichannel signal recovery from non-uniform samples, this problem is not well studied from a theoretical perspective. However, the SLR framework can still be adapted to work well through the synergistic combination of multiple  signal priors (e.g., sparsity and multichannel annihilation priors), achieved by composite lifting \cite{mani2017multi}. More theoretical work is needed on this front to improve the understanding of this problem.}


\textbf{Theoretical guarantees and noise performance:}
{Most of the above analysis assumes that the measurements are noise-free  and the underlying SLR matrix is low-rank. For example, the multichannel annihilation relationship in \eqref{fourierannihilate} is valid only for the noiseless  $k$-space measurements. The model mismatch introduces a slow decay of the singular value spectrum of the Hankel matrix. The above theoretical results are extended to scenarios involving measurement noise in specific SLR models in \cite{ye2016compressive,ongie2017convex}. However, note that SLR formulations can be obtained from many different redundancies within the signal \cite{shin2014calibrationless,haldar2014low,mani2017multi,arvindB0}; the generalization of the theoretical robustness results in \cite{ye2016compressive,ongie2017convex} to general SLR schemes deserves further investigation.}

{
\textbf{Links between machine learning and SLR:}
 Although the aforementioned link between the SLR and deep learning is interesting, there are still  open questions, and more theoretical work is necessary to understand this connection. 
 {The generalizability, optimization landscape, and expressivity (see the extensive list of  references in \cite{ye2019understanding}) 
 in the context of machine-learning based reconstruction are not still
 well-understood, which may be an important research direction}.
}

\vspace{-1em}\section*{Acknowledgement}
JCY is supported by National Research Foundation of Korea under Grant NRF-2016R1A2B3008104. MJ is supported by grants NIH 1R01EB019961-01A1 and R01 EB019961-02S1.

\end{document}